\documentclass[preprint,12pt]{elsarticle}

\usepackage{amssymb}
\usepackage{amsmath}
\usepackage{physics}
\usepackage{float}
\usepackage{hyperref}
\usepackage{comment}
\usepackage{xcolor}
\usepackage{physics}

\usepackage{array}

\newcommand{\textcomment}[1]{}

\newcommand{\Koop}{\boldsymbol{\mathcal{K}}}
\newcommand*{\mbf}[1]{\mathbf{#1}}
\newcommand{\Mcal}{\mathcal{M}}
\newcommand{\Mbar}{\overline{\mathcal{M}}}

\newcommand{\LED}{\textit{LED} }
\newcommand{\iLED}{\textit{iLED} }

\journal{}

\begin{document}

\begin{frontmatter}



\title{Interpretable learning of effective dynamics for multiscale systems}


\author[TAU,IRT]{Emmanuel Menier}
\ead{emmanuel.menier@inria.fr}
\affiliation[TAU]{organization={Tau, Inria / LISN, Université Paris-Saclay / CNRS},
            addressline={1 Rue Raimond Castaing}, 
            city={Gif-sur-Yvette},
            postcode={91190},
            country={France}}
            
\affiliation[IRT]{organization={IRT SystemX},
            addressline={8 avenue de la Vauve}, 
            city={Palaiseau},
            postcode={91120},
            country={France}}

\author[HA,ETH]{Sebastian Kaltenbach}
\ead{skaltenbach@seas.harvard.edu}
\cortext[cor1]{Corresponding author}
\author[IRT]{Mouadh Yagoubi}
\ead{mouadh.yagoubi@irt-systemx.fr}
\author[TAU]{Marc Schoenauer}
\ead{marc.schoenauer@inria.fr}
\author[HA]{Petros Koumoutsakos\corref{cor1}}
\ead{petros@seas.harvard.edu}

\affiliation[HA]{organization={School of Engineering and Applied Sciences, Harvard University},
            addressline={29 Oxford Street}, 
            city={Cambridge},
            postcode={02138}, 
            state={MA},
            country={USA}}

\affiliation[ETH]{organization={Computational Science and Engineering Laboratory, ETH Zürich},
            postcode={CH-8092}, 
            country={Switzerland}}

\begin{abstract}
The modeling and simulation of high-dimensional multiscale systems is a critical challenge  across all areas of  science and engineering. It is broadly believed that even with today's computer advances resolving all  spatiotemporal scales described by the governing equations remains a remote traget. This realization has prompted intense efforts  to develop  model order reduction techniques. In recent years, techniques based on deep recurrent neural networks have produced promising results for the modeling and simulation of complex spatiotemporal systems and offer large flexibility in model development as they can incorporate experimental and computational data. However, neural networks lack interpretability, which limits their utility and generalizability across complex systems. Here we propose a novel framework of Interpretable Learning Effective Dynamics (\iLED) that offers comparable accuracy to state-of-the-art recurrent neural network-based approaches while providing the added benefit of interpretability. The \iLED framework is motivated by Mori-Zwanzig and Koopman operator theory, which justifies the choice of the specific architecture. We demonstrate the effectiveness of the proposed framework in simulations of three benchmark  multiscale systems. Our results show that the \iLED framework can generate accurate predictions and obtain interpretable dynamics, making it a promising approach for solving high-dimensional multiscale systems.
\end{abstract}


\end{frontmatter}



\section{Introduction}

Reliable prediction of critical phenomena, such as weather and  epidemics, depends on the efficiency and veracity of numerical simulations. A vast number of these simulations are founded on models described by Partial Differential Equations (PDEs) expressing multiphysics and multiscale dynamics. Examples include  turbulence \cite{wilcox1988multiscale}, neuroscience \cite{dura2019netpyne}, climate \cite{climatenas} and  ocean dynamics \cite{mahadevan2016impact}. Today we benefit from remarkable efforts in numerical methods, algorithms, software, and hardware and witness simulation frontiers that were unimaginable a couple of decades ago.  However, it is becoming evident that the reliability and energy costs of simulations are reaching their limits \cite{Palmer2015}. Multitudes of spatiotemporal scales in such  systems present high, if not insurmountable, barriers for classical numerical methods. 
Multiscale methods, aim to to resolve this issue by  judicious approximations of the various scales and the interactions between physical processes occurring over different scales. 

Over the years, a number of potent frameworks have been proposed including the equation-free framework (EFF) ~\cite{kevrekidis2004equation, kevrekidis2003equation,bar2019learning}, 
the Heterogeneous Multiscale Method (HMM)~\cite{weinan2003heterognous,weinan2007heterogeneous} and the FLow AVeraged integatoR (FLAVOR)~\cite{tao2010nonintrusive}. 
Their success depends on the separation of scales in  the system dynamics and their capability to capture the transfer of information between scales.  While it is undisputed that  EFF, HMM, FLAVOR have revolutionized the field of multiscale modeling and simulation,  two critical issues  limit their potential:(i) the accuracy of propagating the coarse grained dynamics hinges on the employed time integrators and (ii)ineffective information transfer, in particular from coarse to fine scale dynamics, greatly limits their potential. An alternative and very potent  approach are 
Reduced Order Models (ROMS) \cite{givon_extracting_2004, mezic2005spectral} aim to capture the effective system dynamics and they  can also lead to interpretable descriptions allowing  better understanding of the underlying system \cite{grigo2019physics,kaltenbach2023interpretable}. The success of ROMs hinges on identifying  the low dimensional sub-spaces, on which the dynamics of a system can be summarised.
The existence of such low dimensional sub-spaces has been well established for systems ranging from the 1D Kuramoto-Shivashinsky equation \cite{robinson1994inertial} to highly complex fluid fluid mechanics problems\cite{RowleyReview}. Linear  reduction methods such as the Proper Orthogonal Decomposition, have attracted significant attention as foundations of ROMs\cite{PEHERSTORFER201521} thanks to their simplicity and stability. However  they have shown limitations in the reduction of highly non-linear dynamics \cite{choi2019space} such as those encountered in  turbulent flows, or chemical reactions. Although such methods generally achieve lower performance than non-linear dimensionality reduction due to the aforementioned restriction to a linear mapping, these methods can be used to build analytical reduced-order models based on the chosen POD modes \cite{CD-ROM}. By contrast, neural autoencoders are able to optimally represent the low dimensional, non-linear manifolds on which dynamical systems evolve \cite{lee2020model} and have become the state of the art for non-linear dimensionality reduction. The latent space identified by these autoencoders can then be used in combination with dynamical modeling approaches to construct reduced models of any system of interest. 

More importantly, the deployment of autoencoders for identifying latent space dynamics can be combined with the ideas of EFF,HMM and FLAVOR. This is the core idea of the recently proposed framework of \textit{Learning Effective Dynamics} (\LED) \cite{vlachas2022multiscale}. LED  deploys neural network autoencoders  for dimensionality reduction of the high dimensional samples, thus  identifying the structure of the latent space via a non-linear mapping.  A second type of neural network architecture, the LSTM \cite{hochreiter1997long}, is then deployed to  learn the dynamics of the reduced system. While this framework has shown promising results its functioning remains non-interpretable.
 In this work we present interpretable LED (iLED), that employs for its latent dynamics a theoretically grounded, dynamical model that replaces the LSTM currently in use. The method of iLED is constructed around interpretable linear dynamics and completed by a physically motivated nonlinear closure.
The proposed framework that is closely related to the Mori-Zwanzig \cite{mori1965transport,zwanzig1973nonlinear} and Koopman operator theory \cite{koopman_hamiltonian_1931,mezic2005spectral,mezic2013analysis}.  \\
The use of Mori-Zwanzig and Koopman theories allows our model to have a strong inductive bias which is not the case with black-box models such as Neural ODE and SDE \cite{chen2018neural, li2020scalable}. 

There are a  number of  potent approaches involving linear models for describing the latent dynamics. Two notable cases are models based on   Koopman Operators(KO) \cite{mezic2013analysis,champion2019discovery, lusch2018deep} and Dynamic Mode Decomposition (DMD) \cite{schmid2010dynamic}\cite{williams2015data} .  The KO approaches rely on the assumption  that the latent space dynamics are Markovian and, as such,they may not capture possible memory effects. DMD   can capture certain time dependencies, but it has limitations in systems with strong non-linearities. 
Other approaches that can lead to interpretable solutions \cite{doshi2017towards} include modified RNN architectures \cite{koutnik2014clockwork}, incorporating physical constraints as virtual observables \cite{kaltenbach2020incorporating} or restricting the dynamics of the latent space \cite{kaltenbach_physics-aware_2021-1, kaltenbach2023interpretable}. Moreover, there are some models that employ existing reduced order models and learn a closure term \cite{CD-ROM}\cite{Novati2021}.\\
The iLED, presented in this paper, employs temporal dynamics with explicit linear and non-linear components allowing for  higher  flexibility and interpretability. It is based on both the Mori-Zwanzig formalism \cite{mori1965transport,zwanzig1973nonlinear} as well as KO theory \cite{koopman_hamiltonian_1931,mezic2013analysis}.

The remainder of this paper is structured as follows. In Section \ref{sec:method} we present the general methodological framework with special attention on the connection between our novel framework and the Mori-Zwanzig formalism as well as the Koopman operator theory. Computational aspects related to training the framework and generating predictions are discussed in Section \ref{sec:training}. Afterward, we  present three numerical illustrations  in Section \ref{sec:examples},  and demonstrate the accuracy and interpretability of iLED  over  classical deep learning modeling approaches.
Section \ref{Sec:conlcusions} summarizes the contribution of this work
and suggests directions for future research.

\section{Methodology}
\label{sec:method}

In this section, we present  the  \iLED framework and focus on its connections with the KO theory and the Mori-Zwanzig formalism. 
We  consider high dimensional, potentially non-linear system whose state $\mathbf{\Phi} \in \mathbb{R}^{d_\Phi}$ evolves in time according to an operator $\boldsymbol{F}$:

\begin{equation}
    \dv{\mathbf{\Phi}}{t}=\boldsymbol{F}(t,\mathbf{\Phi})
    \label{eq:FOM}
\end{equation}
This system can result from first principles and may represent the numerical discretization of a PDE such as the Navier-Stokes equations. 
In the following,  we provide the motivation for the choice of neural networks to carry out dimensionality reduction (sec. \ref{sec:DR}), the theoretical justifications for our proposed interpretable reduced dynamics framework (sec. \ref{sec:MZ}) and summarize the  \iLED architecture in section \ref{sec:Arch_ILED}.\\

\subsection{Dimensionality Reduction}\label{sec:DR}

We consider systems where the dimension $d_\Phi$ of the full order system in equation \eqref{eq:FOM} can be exceedingly high compared to the actual intrinsic system dimension. We assume that there  exists a mapping $\mathcal{D}: \mathbb{R}^{d_\Phi} \mapsto \mathbb{R}^{d_z}$,  with $d_z \ll d_\Phi$, such that $\mathbf{\Phi} \approx \mathcal{D}(\boldsymbol{z})$. 
In order to identify and exploit this reduced dimensionality, the mapping $\mathcal{D}$ can be extracted from data using machine learning methods. Classical reduced order modeling methods leverage linear reduction approaches \cite{PEHERSTORFER201521} to construct a basis that builds a matrix $\boldsymbol{V} \in \mathbb{R}^{d_\Phi \times d_z}$ on which both the system's state $\mathbf{\Phi}$ and the dynamics $\boldsymbol{F}$ can be projected~:

\begin{align}
    \boldsymbol{V}^T \mathbf{\Phi} &= \boldsymbol{z},\\
    \dv{\boldsymbol{z}}{t} &= \boldsymbol{V}^T \boldsymbol{F}(t,\boldsymbol{V}\boldsymbol{z}) + \boldsymbol{\epsilon},
\end{align}
where $\boldsymbol{\epsilon}$ is an unknown error term. These linear reduction approaches have the important advantage of being interpretable as they are able to retain parts of the original model $\boldsymbol{F}$. Despite numerous successes with ROMs\cite{Amsallem2011,GEELEN2023115717}, linear reduction can be  inefficient, in terms of dimensionality reduction when compared to non-linear approaches such as neural networks. In many dynamical systems their effective  reduced-order dynamics evolve on strongly non-linear manifolds\cite{vlachas2022multiscale}. The use of neural autoencoders to learn these reduced manifolds allows for a higher degree of reduction, as well as better reconstruction accuracy. Thus, we propose to learn two parameterized non-linear mappings, a decoder $\mathcal{D}(\cdot;\boldsymbol{\theta}_\mathcal{D})$ and an encoder $\mathcal{E}(\cdot;\boldsymbol{\theta}_\mathcal{E})$, such that~:

\begin{align}
    \mathbf{\Phi} &= \mathcal{D}(\boldsymbol{z};\boldsymbol{\theta}_\mathcal{D}),\\
    \boldsymbol{z} &= \mathcal{E}(\mathbf{\Phi};\boldsymbol{\theta}_\mathcal{E})
\end{align}

where $\boldsymbol{\theta}_\mathcal{D}$ and $\boldsymbol{\theta}_\mathcal{E}$ are the parameters of the decoder and encoder and learned during training of the neural networks.\\
However, using a non-linear encoder/decoder structure, the dynamics of the reduced-order system have to be learned afterward or concurrently, as non-linear dimensionality reduction does not allow for the direct reduction of the original model $\boldsymbol{F}$. Existing works \cite{girin2020dynamical,srivastava2015unsupervised}, and more recently the \LED framework \cite{vlachas2022multiscale}, have demonstrated that these reduced dynamics could be directly learned using recurrent neural networks~:
\begin{equation}
    \boldsymbol{z}_{t+1} = \text{RNN}(\boldsymbol{z}_t,\boldsymbol{h}_t;\boldsymbol{\theta}_{RNN}),
\end{equation}
where $\boldsymbol{h}$ is a memory term and $\boldsymbol{\theta}_{RNN}$ are the parameters of the RNN. At the same time, these models have limited interpretability, and, unlike KO or DMD, they cannot be justified by dynamical systems theory.  

\subsection{Framing \iLED within the Mori-Zwanzig formalism}
\label{sec:MZ}
The \iLED framework is based on both the Mori-Zwanzig formalism \cite{mori1965transport,zwanzig1973nonlinear} as well as KO  theory \cite{koopman_hamiltonian_1931,mezic2013analysis}. We first define the KO which acts on observable functions $\mathbf{g}$ of the state of high-dimensional systems $\mathbf{\Phi}$, before introducing the Generalized Langevin Equation (GLE) for a reduced subset of these observables. We subsequently define an appropriate closure term for the GLE and introduce a neural network architecture.\\

\subsubsection{Koopman Operator Theory and the Generalized Langevin Equation}

The KO can quantify the dynamics of an observable of a high-dimensional system and has been employed extensively within ROMs \cite{brunton2021modern,budivsic2012applied,mezic2013analysis}. For the high-dimensional system $\mathbf{\Phi}(t,\mathbf{\Phi}_0)$, a KO can be used to represent the dynamics of the system instead of Equation \ref{eq:FOM}. 
More specifically, an observable $\text{g}:\mathbb{R}^{d_\Phi} \mapsto \mathbb{R}$ of the system $\mathbf{\Phi}$ is advanced in time by the KO, denoted below as $\Koop_t$ \cite{lin2021datadrivenmz}:

\begin{equation}
    \Koop_t \text{g}(\mathbf{\Phi_0}) = \text{g}(\mathbf{\Phi}(t,\mathbf{\Phi_0}))
    \label{eq:Koop}
\end{equation}
Operator $\Koop_t$ is linear, and potentially infinite dimensional. For practical purposes, its operating space can be separated in an observed subspace ($\mathcal{H}_\mathbf{g}$) defined as the space spanned by a chosen set of $M$ observables $\mathcal{M} = \{\text{g}_i\}_{i=1,\ldots,M}$ and an orthogonal subspace $\mathcal{H}_\mathbf{\Bar{g}}$ for which a set of basis functions $\Mbar = \{\Bar{\text{g}}_i\}_{i=M+1,\ldots,\infty}$ can be constructed so that $\langle \text{g}_i,\Bar{\text{g}}_j \rangle=0$ for all $i\in [1,M], j>M$. The dynamics of observables can then be expressed on the basis defined by the set $\mathcal{M} \cup \Mbar$  \cite{lin2021datadrivenmz}:

\begin{align}
    &\dv{}{t}\begin{bmatrix}
         \mbf{g}_{\Mcal} \\ 
         \mbf{g}_{\Mbar}
         \end{bmatrix}
         =\mbf{L}\left[\begin{array}{c}
         \mbf{g}_{\Mcal} \\ 
         \mbf{g}_{\Mbar}
         \end{array}\right] =\left[\begin{array}{c c}
         \mbf{L}_{\Mcal\Mcal} \,\,
         \mbf{L}_{\Mcal\Mbar} \\ 
         \mbf{L}_{\Mbar\Mcal}\,\,
         \mbf{L}_{\Mbar\Mbar}
         \end{array}\right]
         \left[\begin{array}{c}
         \mbf{g}_{\Mcal}\\ 
         \mbf{g}_{\Mbar}
         \end{array}\right]\label{eq:Liouville}.
\end{align}

where $\mathbf{L}$ is a linear operator that corresponds to the infinitesimal generator of $\Koop_t$, $\mathbf{g}_\mathcal{M} = [\text{g}_1, \text{g}_2,\ldots,\text{g}_M]$ are the chosen observables and $\mathbf{g}_{\Mbar} = [\Bar{\text{g}}_{M+1}, \Bar{\text{g}}_{M+2},\ldots,\Bar{\text{g}}_{\infty}]$ are the orthogonal observables. Note that the operator $\mathbf{L}$ is separated in four parts, with $\mbf{L}_{\Mcal\Mcal}$ the dynamics in the observed subspace, $\mbf{L}_{\Mbar\Mbar}$ the orthogonal dynamics and $\mbf{L}_{\Mbar\Mcal}$ and $\mbf{L}_{\Mcal\Mbar}$ the exchanges between the observed and orthogonal subspaces.

The above system can be solved for $\mathbf{g}_{\Mbar}$ as follows:

\begin{equation}
    \mbf{g}_{\Mbar} (t) = \int_0^t e^{(t-s) \mbf{L}_{\Mbar\Mbar}} \mbf{L}_{\Mbar\Mcal} \mbf{g}_{\Mcal} (s) ds + e^{t \mbf{L}_{\Mbar\Mbar}} \mbf{g}_{\Mbar} (0), \qquad t>0 \label{eq:small_scales_dyn}.
\end{equation}

Finally, using Eq.\eqref{eq:small_scales_dyn} in \eqref{eq:Liouville}, an expression for the dynamics of the observables $\mathbf{g}_{\Mcal}$ is obtained:

\begin{equation}
    \dv{\mbf{g}_{\Mcal}}{t} =
    \mbf{L}_{\Mcal\Mcal} \mbf{g}_{\Mcal} +
    \mbf{L}_{\Mcal\Mbar} \int_0^t e^{(t-s) \mbf{L}_{\Mbar\Mbar}} \mbf{L}_{\Mbar\Mcal} \mbf{g}_{\Mcal} (s) ds + \mbf{L}_{\Mcal\Mbar} e^{t \mbf{L}_{\Mbar\Mbar}} \mbf{g}_{\Mbar} (0).
    \label{eq:GLE}
\end{equation}

The above expression describes the dynamics of the partially observed state of a system and has the same form as the Generalized Langevin Equation derived in the Mori Zwanzig formalism. It still depends on the unobserved part of the initial condition ($\mathbf{g}_{\Mbar}(0)$) via the last term and thus is not a closed equation for $\mbf{g}_{\Mcal}$ only. However, this last term is often modeled as noise or simply ignored in several modeling approaches \cite{chorin2007problem,darve2009computing,kondrashov2015data,lin2021datadrivenmz}. In the following section the conditions under which this term can be accounted for are explained in more detail.

\subsubsection{Closing the GLE}
\label{sec:ClosingGLE}
The last term in Eq.\eqref{eq:GLE} depends on information that is unavailable as it is orthogonal to the observed subspace. However, this term vanishes if the history of the observed subspace is known, and the orthogonal dynamics ($\mathbf{L}_{\Mbar\Mbar}$) are dissipative. Indeed, if the history of the system is known, we can re-write Eq.\eqref{eq:small_scales_dyn} for any initial condition $[\mathbf{g}_{\Mcal}(-\tau),\mathbf{g}_{\Mbar}(-\tau)], \tau>0$:

\begin{equation}
    \mbf{g}_{\Mbar} (t) = \int_{-\tau}^t e^{(t-s) \mbf{L}_{\Mbar\Mbar}} \mbf{L}_{\Mbar\Mcal} \mbf{g}_{\Mcal} (s) ds + e^{(t+\tau) \mbf{L}_{\Mbar\Mbar}} \mbf{g}_{\Mbar} (-\tau).\label{eq:small_scales_dyn_simpl}
\end{equation}

The last term in Eq.\ref{eq:small_scales_dyn_simpl} vanishes for $\tau \to \infty$, if the orthogonal dynamics $\mathbf{L}_{\Mbar\Mbar}$ are dissipative, which is often a reasonable assumption as, for instance, when the orthogonal (unobserved) subspace corresponds to the small scales of a dynamical system. Under such hypothesis, we obtain the following closed equation for the dynamics of the observed subspace:

\begin{align}
    &\mbf{g}_{\Mbar} (t) = \int_{-\infty}^t e^{(t-s) \mbf{L}_{\Mbar\Mbar}} \mbf{L}_{\Mbar\Mcal} \mbf{g}_{\Mcal} (s) ds,\\
    \implies &\dv{\mbf{g}_{\Mcal}}{t} =
    \mbf{L}_{\Mcal\Mcal} \mbf{g}_{\Mcal} +
    \mbf{L}_{\Mcal\Mbar} \int_{-\infty}^t e^{(t-s) \mbf{L}_{\Mbar\Mbar}} \mbf{L}_{\Mbar\Mcal} \mbf{g}_{\Mcal} (s) ds.\label{eq:ClosedGLE}
\end{align} 

In the following subsection, we show that the various operators expressed in the closed GLE (Eq. \eqref{eq:ClosedGLE}) can be learned from data to derive an interpretable and theoretically sound model for the reduced dynamics of physical systems.

\subsection{The \iLED architecture}
\label{sec:Arch_ILED}

To construct the \iLED architecture, we first identify the observables $\mathbf{g}_{\Mcal}$ with the learned subspace of the neural encoder $\mathcal{E}$ so that $\mathbf{g}_{\Mcal} \equiv \boldsymbol{z} = \mathcal{E}(\mathbf{\Phi};\boldsymbol{\theta}_\mathcal{E})$. We then learn the various operators $L_{xx}$ that express the different parts of the Mori-Zwanzig formalism in equation \eqref{eq:ClosedGLE}. 

The observed dynamics $\mbf{L}_{\Mcal\Mcal}$ can be directly learned as a linear operator, denoted $A_\theta \in \mathbb{R}^{d_z \times d_z}$ below. However, because both operators $\mbf{L}_{\Mbar\Mcal}$ and $\mbf{L}_{\Mcal\Mbar}$ are possibly infinite dimensional, they need to be approximated. We propose to learn these operators as non-linear transformations of the observables $\boldsymbol{z}$. Justifications for this choice will be detailed in section \ref{sec:Remarks}. 

\begin{figure}
    \centering
    \includegraphics[trim={6cm 0 0 0},clip,scale=0.25]{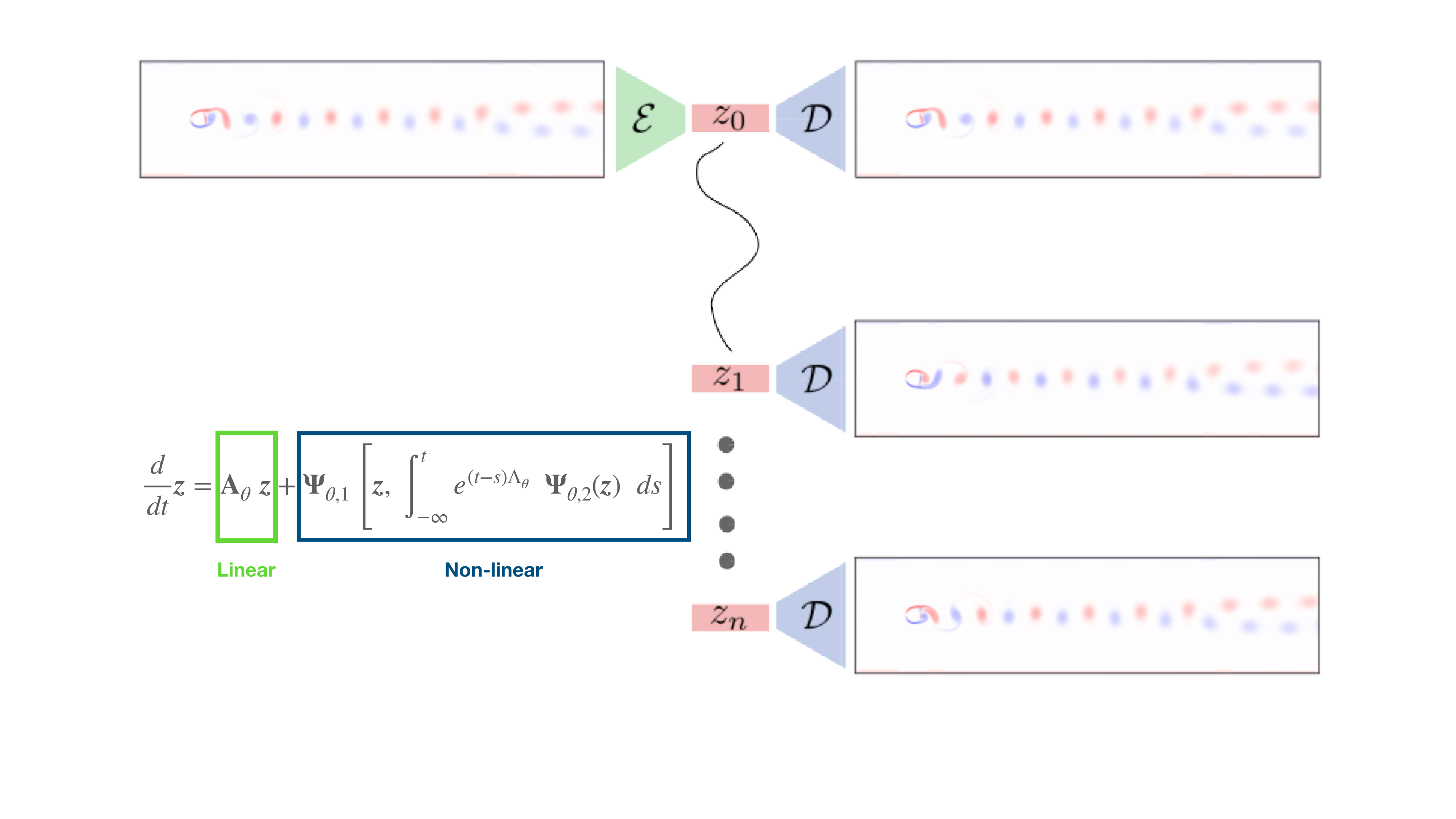}
    \caption{ILED architecture: The high-dimensional system is encoded to a lower-dimensional representation using the encoder $\mathcal{E}$. The lower-dimensional representation is propagated in time using a \textcolor{green}{linear} and a \textcolor{blue}{non-linear} part based on the Mori-Zwanzig formalism. With the help of a decoder $\mathcal{D}$, the high-dimensional system is subsequently reconstructed.}
    \label{fig:iled_schematric}
\end{figure}

We introduce two neural networks $\mathbf{\Psi}_1(\cdot;\theta) : \mathbb{R}^{d_h+d_z} \mapsto \mathbb{R}^{d_z}$ and $\mathbf{\Psi}_2(\cdot;\theta) : \mathbb{R}^{d_z} \mapsto \mathbb{R}^{d_h}$, where $d_h$ is a user-defined parameter, and model the orthogonal dynamics $\mbf{L}_{\Mbar\Mbar}$ as a negative diagonal operator $\Lambda_\theta \in \mathbb{R}^{d_h \times d_h}_{-}$. This choice is consistent with the assumption that the orthogonal dynamics are dissipative and significantly simplifies certain computations such as the initialization of the non-markovian (or memory) term in the model. This leads to the \iLED architecture in Figure \ref{fig:iled_schematric}.

The key part of this novel architecture is the temporal dynamics of the \iLED state $\boldsymbol{z}$:
\begin{equation}\label{eq:iLED}
    \dv{}{t}\boldsymbol{z} = \mathbf{A}_\theta \boldsymbol{z} + \mathbf{\Psi}_{\theta,1} \left[\boldsymbol{z}, \int_{-\infty}^t e^{(t-s)\Lambda_\theta} \mathbf{\Psi}_{\theta,2} (\boldsymbol{z})  ds\right]
\end{equation}
We note that the operator $\Psi_1$ now also takes as argument the reduced state $\boldsymbol{z}$ itself. This choice ensures that the nonlinear part of the \iLED model has access to the state at the current time. In terms of the framework derived above this corresponds to the network $\mathbf{\Psi}_2$ learning an identity of the state in part of its output so that $\mathbf{\Psi}_2(\boldsymbol{z}) = [\mathbf{z},\ldots]$ and the corresponding entries of the diagonal matrix $\Lambda$ going to $-\infty$. Indeed, the integral in Eq.\eqref{eq:iLED} is a low-pass filtering of the trajectories of each dimension of $\mathbf{\Psi}_2(\boldsymbol{z})$ with a cutoff frequency of $\frac{-1}{\lambda_i}$ with $\lambda_i$ the $i_{\textit{th}}$ entry in the diagonal of $\mathbf{\Lambda}$. Thus, $\lambda_i$ going to negative infinity implies that no frequency is filtered in the trajectory, which is equivalent to directly considering the state at the current time.\\ This proposed  \iLED architecture allows us to directly learn the various terms of the Mori-Zwanzig formalism from data. The details on the training strategy are  given in section \ref{sec:training}. First, we provide additional justifications for the model.

\textcomment{
where $\mathbf{A}_\theta$ is a trainable linear operator, $\Lambda$ is a negative definite matrix, $\mathbf{\Psi}_{\theta,1}$ is a neural network trained to close the dynamics of the observables and $\mathbf{\Psi}_{\theta,2}$ is a neural network trained to lift the observables in a higher dimensional space.

The parallel between the iLED (Eq.\eqref{eq:LID})  formulation and the closed GLE (Eq.\eqref{eq:CloseGLE}) can be used to motivate the modeling choices:

\begin{itemize}
    \item The operator $\mathbf{A}_\theta$ can be directly identified with the observable dynamics $\mathbf{L}_{\Mcal\Mcal}$. 
    \item The choice of a negative definite matrix $\Lambda$ is consistent with the hypothesis that the orthogonal dynamics $\mathbf{L}_{\Mbar\Mbar}$ are dissipative.
\end{itemize}

We note that the  link between the two equations can not be formally established as the linear operators $\mathbf{L}_{\Mbar\Mcal}$, which projects the observables in the orthogonal (infinite) subspace, and $\mathbf{L}_{\Mcal\Mbar}$ which projects the dynamics closure in the observed subspace, are modeled using finite non-linear operators. This particular choice of describing an infinite linear operator using a finite non-linear model is analogous to the Koopman idea of modeling non-linear dynamics using an infinite linear operator but the theoretical justification for this choice remains to be established.

}

\subsubsection{Remarks on the Approximation of an infinite linear operator}\label{sec:Remarks}
 Deep Neural networks are universal approximators for non-linear operators when both the given input and output of the operator are compact \cite{chen1995universal}.  This property has been successfully employed in appropaches such as the Deep Operator Network \cite{lu2021learning,kissas2022learning} and Fourier Neural Operators (FNOs) \cite{li2021fourier}. However here, the subspace is infinite, and thus not compact. The Koopman operator, our starting point in Equation \ref{eq:Koop}, is generally represented by a finite-dimensional operator with reasonable accuracy. In fact, this is a key assumption for all main data-driven Koopman models\cite{bollt2017edmd,otto2019lran,brunton2021modern}. We assume that the same assumption holds here, and thus are dealing with a finite-dimensional orthogonal space whose operators can be approximated by neural networks.\\


\subsection{Training the \iLED architecure}
\label{sec:training}


A key difficulty in the present methods is the choice of the latent dimension $d_z$. The best choice is a dimension close to the intrinsic dimension of the problem at hand. In the common case where it is unknown, several approaches can be used to select this parameter. One is to directly apply hyperparameter optimization approaches such as grid search to the problem. \textit{i.e.}, train autoencoders with increasing latent dimensions and select the dimension when the reconstruction performance of the autoencoder starts to plateau. However, this approach can be expensive when applied to problems that use high-dimensional representations such as  fluid flows. To avoid these expensive computations,  statistical analysis such as estimation of the  correlation dimension \cite{nolitsa}  or of the fractal dimension\cite{Falconner2003fractal} can be employed to approximate the dimension of the attractor.\\

Having chosen this latent dimension $d_z$, we can set up the model for the latent dynamics.
We first  re-arrange the integro-differential equation \eqref{eq:iLED} into a coupled system of ordinary differential equations. First, we define an intermediate term $\boldsymbol{h}$~:

\begin{equation}
    \boldsymbol{h}(t) = \int_{-\infty}^t e^{(t-s)\Lambda_\theta} \mathbf{\Psi}_{\theta,2} (\boldsymbol{z})  ds.
\end{equation}
This $\boldsymbol{h}$ term corresponds to the \textit{memory} of the model, which can be advanced in time in parallel of the reduced order state as follows~:
\begin{align}
\label{eq:iLED_ODE}
\begin{split}
    \dv{\boldsymbol{z}}{t} &= \mbf{A}_\theta \boldsymbol{z} + \mbf{\Psi}_{\theta,1}(\boldsymbol{z},\boldsymbol{h}),\\
    \dv{\boldsymbol{h}}{t} &= \mbf{\Psi}_{\theta,2}(\boldsymbol{z}) + \Lambda_\theta \boldsymbol{h}.
\end{split}
\end{align}

With this time-continuous architecture, \iLED  can be used in combination with any standard ODE integrator. In this work, we used the semi-implicit Runge-Kutta (siRK) scheme \cite{kar2006sirk3} to advance the \iLED state $[\boldsymbol{z},\boldsymbol{h}]$ in time. This scheme takes advantage of dynamics that efficiently separate a linear and a non-linear part and is very efficient for the simulation of stiff dynamics,. The latter feature is important for  \iLED as the dynamics can be stiff and unstable before being fully trained. The corresponding equations can be found in \ref{sec:AppB}\\

The siRK integration scheme is used in combination with the adjoint scheme for back-propagating through an ODE \cite{chen2018neural}  to train the \iLED architecture. We train the model in an \textit{end-to-end} fashion, that is to say, both the neural autoencoder $\{\mathcal{E},\mathcal{D}\}$ and the dynamics are optimized simultaneously,  using the combined loss~:

\begin{equation}\label{eq:Combined_Loss}
    \mathcal{L} = \mathcal{L}_\text{rec} + \alpha \mathcal{L}_\text{forecast}.
\end{equation}

where $\mathcal{L}_\text{rec}$ and $\mathcal{L}_\text{forecast}$ are respectively the reconstruction and forecast losses, and $\alpha$ controls their relative importance.

The reconstruction loss $\mathcal{L}_\text{rec}$ drives the autoencoder to accurately reconstruct the true full order trajectory $\Phi^\star_{t_i}$~:

\begin{equation}
    \mathcal{L}_\text{rec} = \frac{1}{N_t} \sum_{i=1}^{N_t} \Vert \Phi^\star_{t_i} - \mathcal{D}(\mathcal{E}(\Phi^\star_{t_i})) \Vert_2^2.
\end{equation}

The forecast loss $\mathcal{L}_\text{forecast}$ pushes the model to accurately predict the reduced state $z$~:
\begin{align}
\begin{split}
    \mathcal{L}_\text{forecast} &= \frac{1}{N_t} \sum_{i=1}^{N_t} \Vert \hat{\boldsymbol{z}}_{t_i} - \mathcal{E}(\Phi^\star_{t_i}) \Vert_2^2,
\end{split}
\end{align}
where $\hat{\boldsymbol{z}}$ is calculated according to Equation \ref{eq:iLED_ODE}.

This aggregated loss is sufficient to train the \iLED architecture. However, additional terms can be added to improve performance. \ref{sec:Loss} details the various additions that were used to obtain the results presented in this work, and \ref{sec:eng_details} provides more details regarding other design choices such as the initialization for the memory term $\boldsymbol{h}$. The next section presents the numerical experiments carried out to demonstrate the abilities of the \iLED method.

\section{Numerical Experiments}
\label{sec:examples}
The capabilities of \iLED are demonstrated on three benchmark problems: The FitzHugh-Nagomo model, a simple 1D equation with periodic dynamics; The chaotic dynamics presented by the Kuramoto-Shivasinsky equation; The incompressible Navier-Stokes equations describing flow around a cylinder in two different Reynolds numbers ($100$ and $750$).

\subsection{Example 1: The FitzHugh-Nagomo Model}\label{sec:FHN}

The FitzHugh-Nagomo model \cite{rocsoreanu2012fitzhugh} has been extensively used in biology, physics and neuroscience for the study of the  dynamics of excitable systems. The model consists of two  coupled PDEs that describe the dynamics of a fast-acting variable $u(x,t) \in \mathbb{R}, x \in \Omega = [0,L], t \in [0,T]$, inhibited by a slower variable $v(x,t) \in \mathbb{R}$~:

\begin{align}
    \frac{\partial u}{\partial t} &= D_u \frac{\partial^2 u}{\partial x^2} + u - u^3 -v,\\
    \frac{\partial v}{\partial t} &= D_v \frac{\partial^2 v}{\partial x^2} + \epsilon \left( u -\alpha_1 v - \alpha_0\right).
\end{align}
 This separation of time scales is controlled by parameter $\epsilon$, set here to $\epsilon = 0.006$. The other model parameters are chosen as follows : $D_u = 1,\, D_v=4,\, L=20,\, \alpha_0=-0.03$ and $\alpha_1 = 2$, to replicate the experiment presented in \cite{vlachas2022multiscale}.
The computational domain $\Omega$ is discretized using a grid of $N=101$ points. The problem is solved starting from 5 different initial conditions using the Latice-Boltzmann method \cite{LB} and its implementation is provided in \cite{vlachas2022multiscale}. The data is sampled at rate $\Delta t =1 s$ to obtain 5 trajectories of $451$ seconds each.  Two of those trajectories are set aside for validation and the others are used for training. An additional trajectory of $10^4$ seconds is simulated for testing purposes.

\begin{figure}[h]
    \centering
    \includegraphics[width=\textwidth]{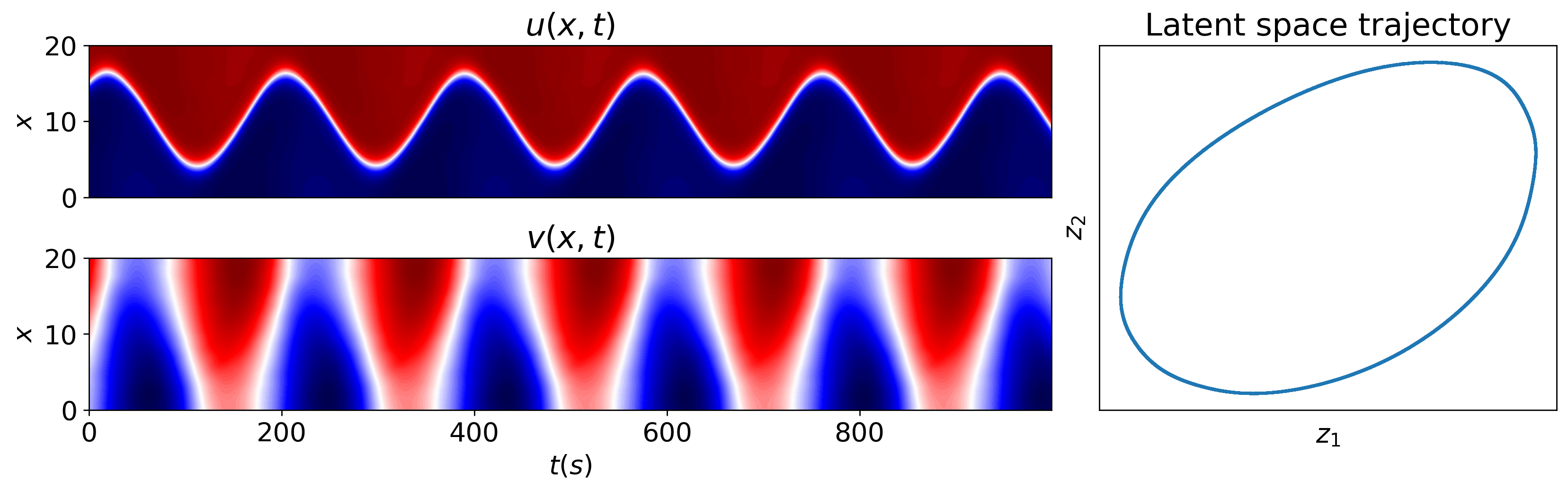}
    \caption{Visualization of the FHN model's dynamics. The evolution of the full state for a subset of the test trajectory is presented on the left. The right hand side of the plot displays the latent manifold learned by an autoencoder using latent dimension $d_z=2$.}
    \label{fig:FHN_intro}
\end{figure}

By training various autoencoders to reconstruct the training trajectories described above, we determined that the optimal latent dimension was $d_z = 2$, as the reconstruction accuracy evaluated from the validation trajectories saturates for higher dimensions. This result is consistent with the oscillatory nature of the dynamics and highlights the efficiency of non-linear dimensionality reduction. Indeed, a linear method such as PCA requires up to $16$ latent dimension (see \cite{vlachas2022multiscale} \textit{figure 2-A}) to achieve the same level of accuracy. A visualization of the system evolution, as well as the corresponding latent trajectory are presented in Figure \ref{fig:FHN_intro}.

\begin{figure}[h]
    \centering
    \includegraphics[width=\textwidth]{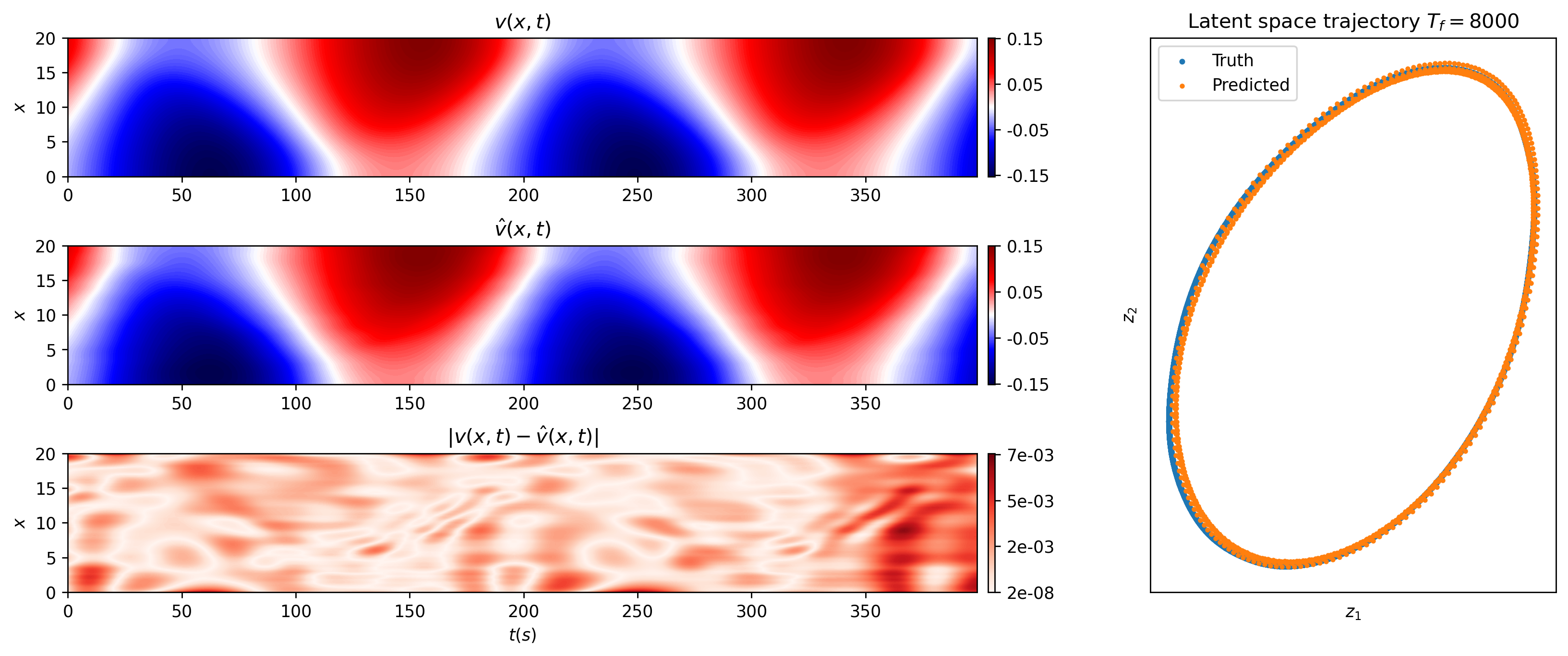}
    \caption{Forecasting performance of the \iLED method on the FHN case. From top to bottom: true inhibitor field, predicted inhibitor field and absolute error between the two. The right hand side presents the true and predicted latent trajectory for an integration period of $8000 s$. \textit{NB :} only the inhibitor $v$ field is presented for clarity, as it is harder to predict than the activator field $u$.}
    \label{fig:FHN_Forecast}
\end{figure}

An \iLED dynamical model is also trained at the same time as the autoencoder, using the procedure described in Section \ref{sec:training} (the hyperparameters used are detailed in \ref{sec:params_fhn}). Figure \ref{fig:FHN_Forecast} presents the results obtained by simulating the final model on the test trajectory. The Figure shows that the \iLED model is able to accurately reconstruct the full order system state from the latent code $z$. Moreover, the dynamics is accurately captured: the model remains on the true latent attractor even after a very long integration.

\begin{figure}[h]
    \centering
    \includegraphics[width=\textwidth]{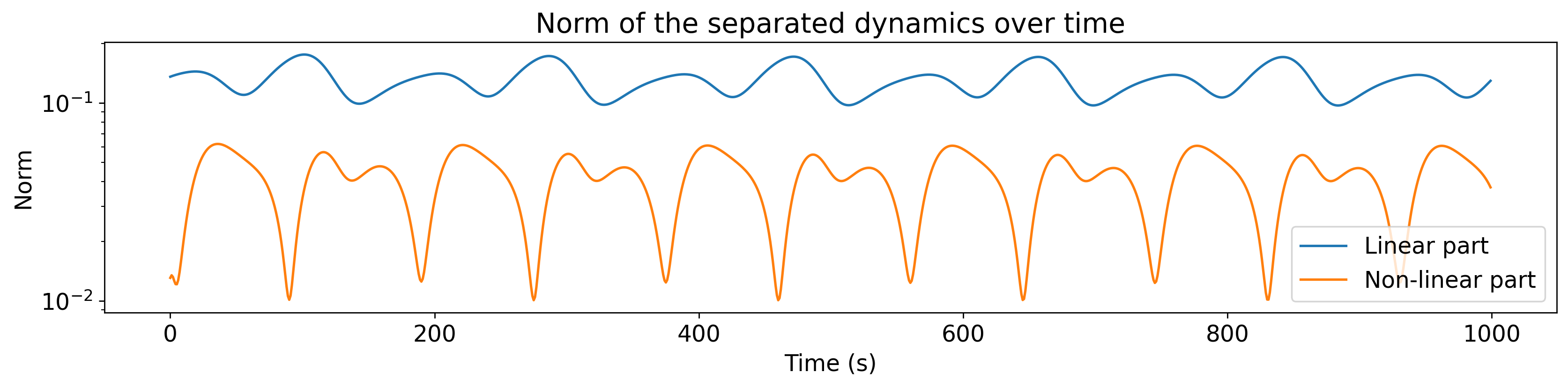}
    \caption{Norm of the dynamics parts}
    \label{fig:FHN_Norm}
\end{figure}

The  \iLED  is particularly well-suited for this case, and highly interpretable. Due to the optimal latent dimension $d_z = 2$, the linear part of the \iLED dynamics exhibits a single natural frequency, aligning with the periodic nature of the dynamics. The learned frequency is approximately $5.74 mHz$, while the primary frequency extracted from the true system data using a Fourier Transform is $5.37 mHz$. This comparison demonstrates that the operator has accurately captured the driving frequency of the system, allowing the linear part of the \iLED model to support most of the dynamics. A close examination of the norm of the dynamics separately for the linear and non-linear terms (Figure \ref{fig:FHN_Norm}) confirms this result: The figure clearly shows that the dynamics is mainly supported by the linear term, the contribution of the nonlinear term being approximately one order of magnitude smaller.  It is important to note that the nonlinear term still plays a role in this case, as the learned latent attractor is not perfectly circular: A purely linear model would inevitably diverge from the true trajectory.

\newpage
\subsection{Example 2: The Kuramoto-Sivashinsky Equation}\label{sec:KS}

The Kuramoto-Sivashinsky (KS) equation\cite{hyman1986kuramoto}, serves as a model for a broad range  of physical systems. It is a prototypical example of a nonlinear partial differential equation, and exhibits a rich variety of behaviors, including the emergence of self-sustained oscillations, the formation of coherent structures, and the occurrence of spatiotemporal chaos, making it an excellent test-bed for ROMs.

The KS equation can be expressed as:
\begin{align}
\begin{split}
    \frac{\partial u}{\partial t}& + \frac{\partial^2 u}{\partial x^2} + \frac{\partial^4 u}{\partial x^4} + u\frac{\partial u}{\partial x} = 0,\\
    u(x,t)& \in \mathbb{R},\, x \in [0,L],\, t \in [0,T],\\
    u(0,t)& = u(L,t),
\end{split}
\end{align}

where u(x, t) represents the unknown scalar field, and L is the length of the computational domain, that controls the nature of the dynamics. We use here $L=22$, a common value for the study of this problem (\cite{vlachas2022multiscale,alessandro2022control}) which yields a dynamical system that evolves on a stable attractor with a characteristic dimension approximately equal to $8$ (a higher dimensional attractor than the attractor of the  FitzHugh-Nagomo model studied in previous Section). Moreover, the KS system develops chaotic dynamics under these conditions, which significantly increases the complexity of the learning problem, as small errors naturally compound over time during the simulation.

\begin{figure}[h]
    \centering
    \includegraphics[width=\textwidth]{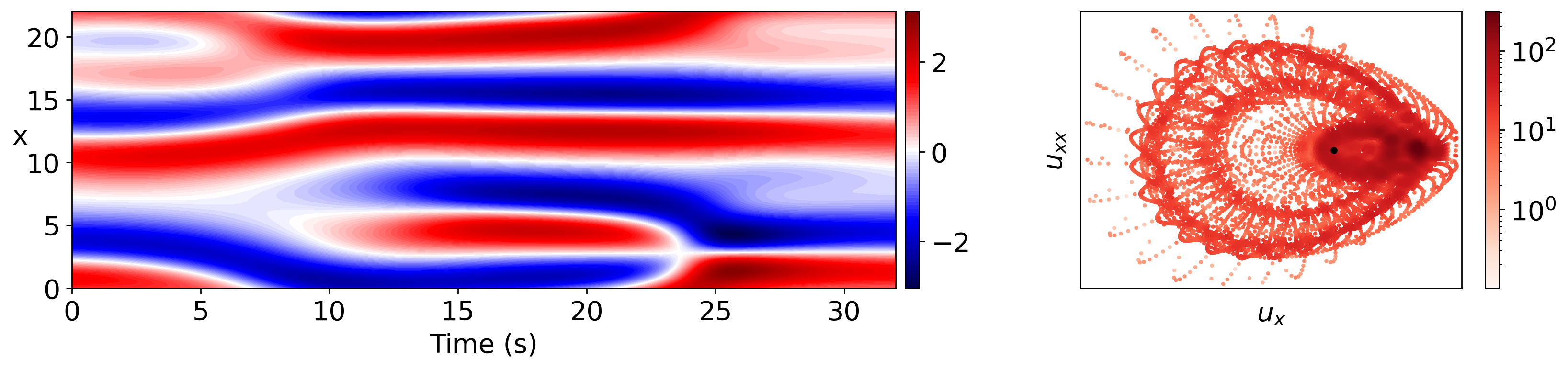}
    \caption{Two views of a training trajectory for the Kuramoto-Shivasinsky case (see text).}
    \label{fig:KS_Intro}
\end{figure}

The equation is discretized on a spectral basis of N=64 Fourier modes, and advanced in time using a Semi implicit Runge-Kutta scheme \cite{kar2006sirk3}. We generate 2048 training trajectories starting from random initial conditions, and 64 others for validation. The initial conditions are all advanced in time for $3000$ "warm-up" steps of length $\delta t = 0.025 s$, which are discarded as they account for the transition from the random initial conditions to the chaotic attractor. The next 1280 steps are then sub-sampled with a $\Delta t = 0.25 s$ in order to obtain the training and validation data. Finally, one hundred new initial conditions are simulated with a longer time horizon ($800 s$) for testing purposes. The evolution of one of the training trajectories is presented in figure \ref{fig:KS_Intro}, as well as a visualization of the joint probability density $p(\frac{\partial u}{\partial x},\frac{\partial^2 u}{\partial x^2})$, which is a helpful way of visualizing the dynamics of the KS equation.

\begin{figure}[h]
    \centering
    \includegraphics[width=\textwidth]{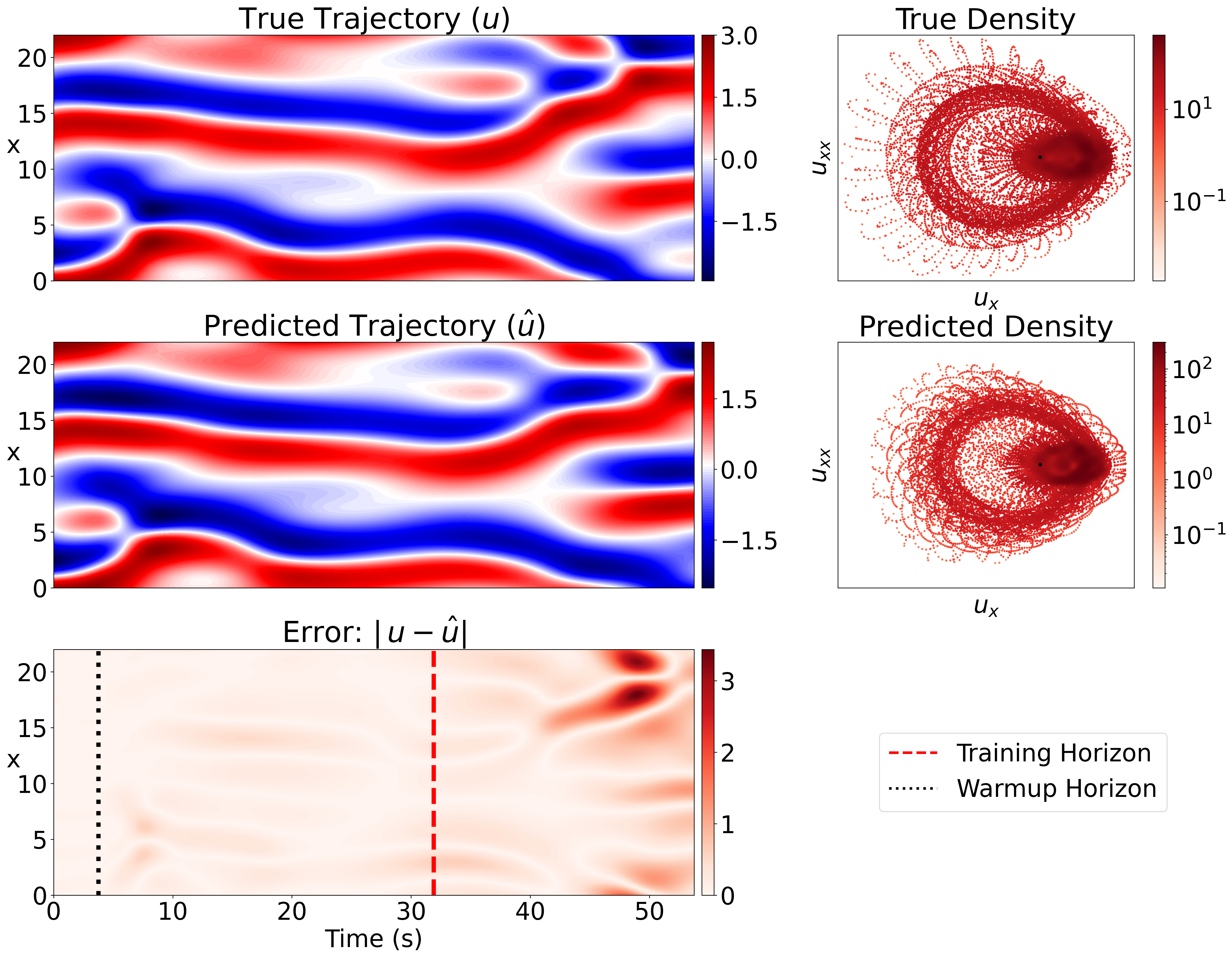}
    \caption{Results obtained with the \iLED method on a test trajectory. \textit{Dashed black line}: Horizon of the warm-up required to initialize the memory of the model, \textit{Dashed red lined}: time horizon used to train the model.}
    \label{fig:KS_forecast}
\end{figure}

Applying the \iLED method
\footnote{Details on the architecture and hyperparameters used can be found in \ref{sec:params_KS}}, we find that the reconstruction performance of the autoencoder used for dimensionality reduction does not improve for latent dimensions greater than  $d_z = 8$. This is  in accordance with the previously discovered  \cite{robinson1994inertial}
intrinsic dimension of the KS attractor. Figure \ref{fig:KS_forecast} presents results obtained on a test trajectory with a trained \iLED model. 

The results in figure\ref{fig:KS_forecast} demonstrate that the \iLED method is able to correctly capture the dynamics of the system on a previously unseen trajectory, and for a time horizon at least as long as its training.As expected,  the forecasting error does increase for longer integration times. The chaotic nature of the problem makes it increasingly hard for a model to accurately follow the true system trajectory. Moreover, we note that despite leaving the true trajectory, the obtained attractor, visualized through the densities of the derivatives, is well captured.

\begin{figure}[h]
    \centering
    \includegraphics[width=\textwidth]{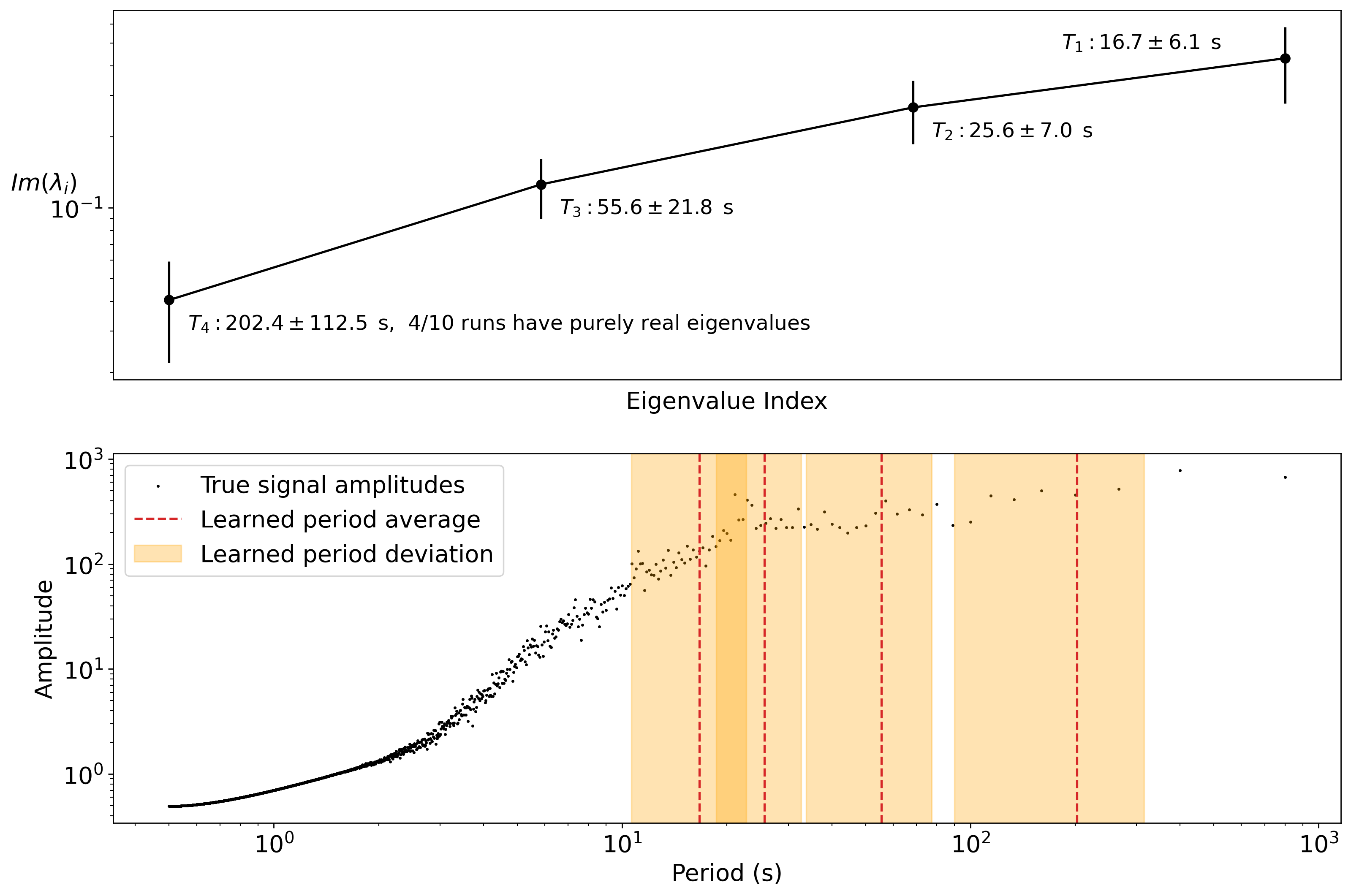}
    \caption{\textit{Top} : Eigenvalues $\lambda_i$ of the \iLED linear operator (average and std. dev. over ten different training runs). \textit{Bottom}: Fourier transform of a test trajectory averaged over the computational domain, the natural periods of the \iLED operator are also displayed for comparison. Note that several runs learned one purely real eigenvalue, meaning that the learned period is infinite, thus not included in the computation of the largest period ($T_4$).}
    \label{fig:KS_eigens}
\end{figure}

We also examine the eigenvalues of the learned linear operator in  \iLED for the FHN case (section \ref{sec:FHN}). Figure \ref{fig:KS_eigens} shows the natural frequencies learned by the \iLED models after training under ten different random seeds. Asthe KS system is not driven by a single main frequency, it is interesting to note that the different model initializations led to learning  a similar range of frequencies. Moreover, the natural frequencies of the \iLED linear operator are coherent with the frequencies observed in the data. Figure \ref{fig:KS_eigens} displays the Fourier transform of a test trajectory, showing that a large range of frequencies is present in the data. The figure also shows that this range is covered by the various frequencies learned by the \iLED linear operator, suggesting that while the chaotic attractor does not directly correspond to a periodic cycle in latent space, this cycle is still relevant to the system dynamics.

\subsection{Example 3: Flow around a two dimensional  circular Cylinder}\label{sec:Cylinder}

Finally, we apply the \iLED method to simulations of the uniform viscous (with viscosity $\nu$) flow with speed $U$ past a 2D circular cylinder, with diameter  $D$, using the Navier Stokes equations. The complexity of this case is controlled by the Reynolds number ($Re = UD/\nu$). We consider  flows with two  different Reynolds numbers: $\textit{Re}=100$, the standard value used to benchmark ROMs\cite{callaham2019robust} and  $\textit{Re}=750$, where  the system then exhibits much more complex dynamics.

\begin{figure}[h]
    \centering
    \includegraphics[width=\textwidth]{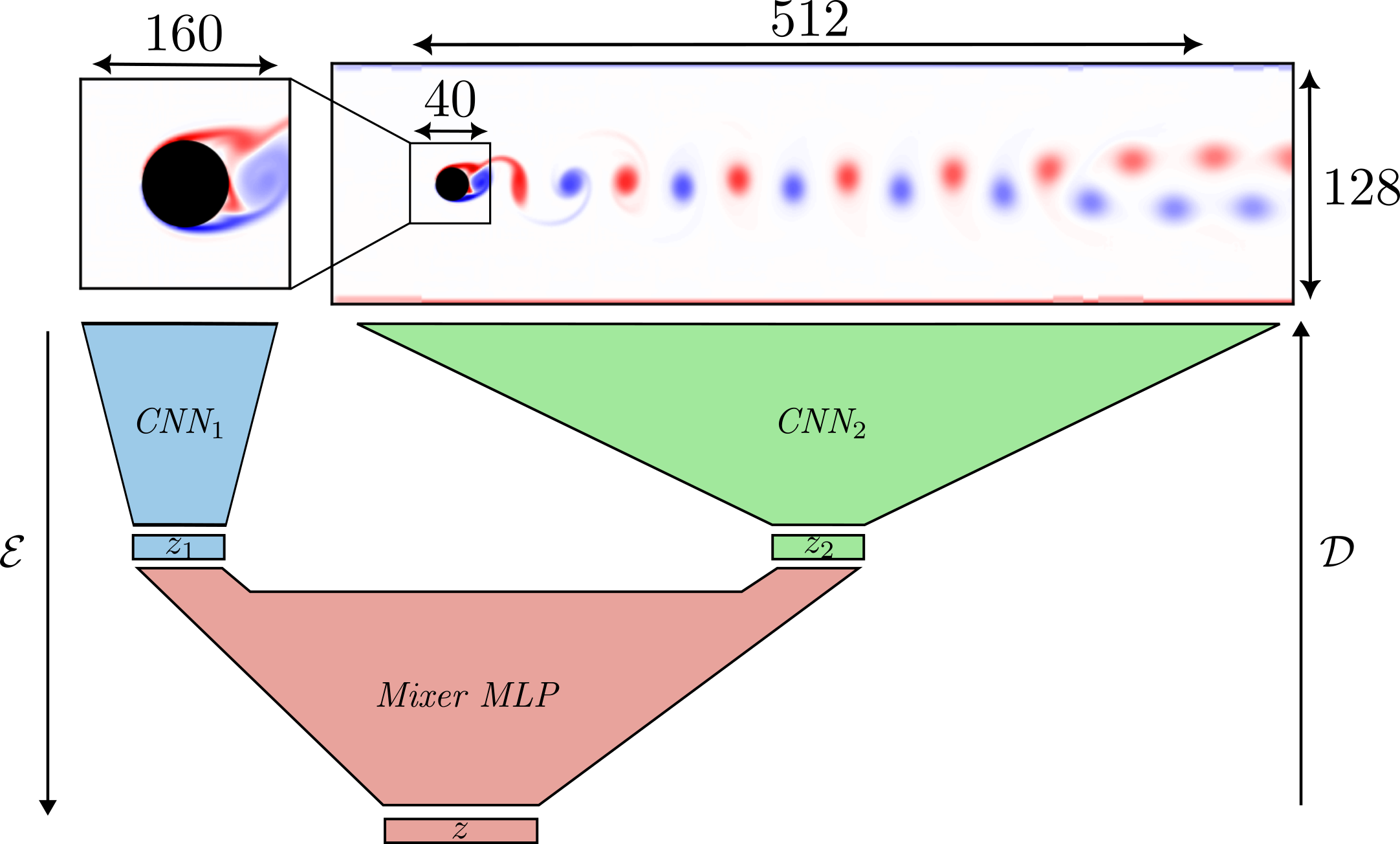}
    \caption{Multiscale architecture used to model the cylinder flow. The area around the cylinder is rendered at four times the resolution of the rest of the field, as it is where the dynamics are most complex.}
    \label{fig:MultiscaleAE}
\end{figure}

In both cases, the incompressible Navier-Stokes equations are solved using an adaptive meshing and time stepping Basilisk solver \citep{popinet2020basilisk}. The generated data is then interpolated on a cartesian grid to ensure compatibility with convolutional neural networks. To construct the autoencoder, we use an approach similar to the one proposed in \cite{AdaLED}, using  a higher resolution around the cylinder to capture the process of vorticity generation. Hence, we deploy separate convolutional encoders with different resolutions for the region around the cylinder and the remaining of the computational domain. These two encoders produce two intermediate latent representations $z_1$ and $z_2$ which are passed through an additional \textit{mixer} multi layer perceptron to compute the latent code $z$. This \textit{mixer} network is used to ensure that each dimension of the latent state $z$ can encode information for both the higher and lower resolution parts of the state, which is important as the \iLED linear operator $\mathbf{A}_\theta$ acts on the full latent state $z$. This multi-scale architecture is illustrated in figure \ref{fig:MultiscaleAE}. Additional details on the architecture and hyperparameters used can be found in \ref{sec:params_flow}

For both Reynolds numbers, the problem is simulated for $100 s$. The first warm-up twenty seconds are discarded as they correspond to the transition from the initial condition. The rest of the trajectory is sub-sampled with a $\Delta_t = 0.02 s$ yielding a trajectory of $4000$ points. The first $2500$ points are used for training, and the last $1500$ are set aside for validation. 

\begin{figure}
    \centering
    \includegraphics[width=\textwidth]{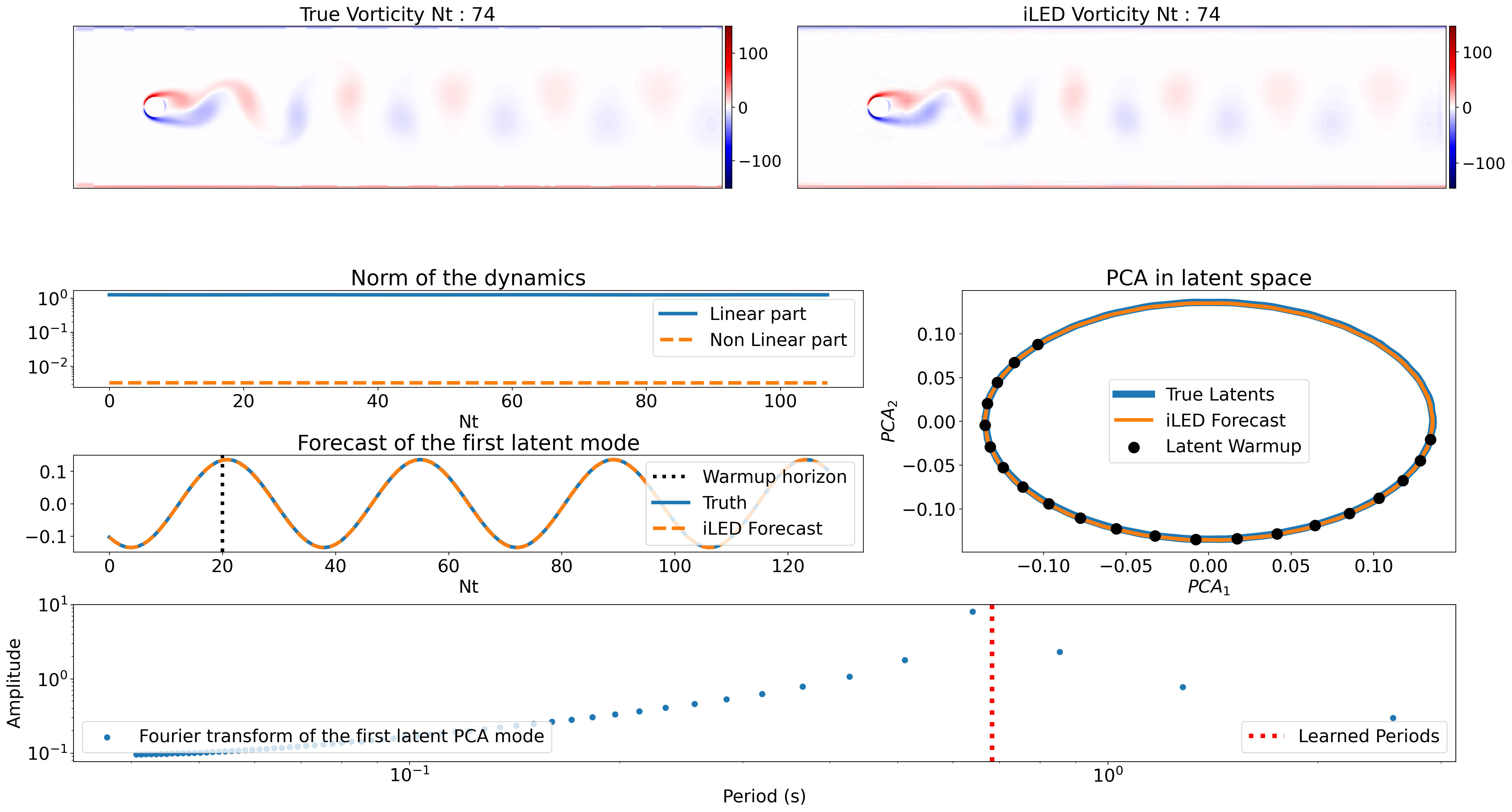}
    \caption{Results obtained with the \iLED method on the case of the cylinder flow at a Reynolds number of 100. }
    \label{fig:Re100_results}
\end{figure}

The results obtained by training an \iLED model for the $\textit{Re} = 100$ case are presented in figure \ref{fig:Re100_results}. We used a latent dimension of $d_z = 3$, which is higher than the minimal dimension $2$ required to represent the limit cycle of the system, but yielded better modeling performance according to the combined loss (Eq.\eqref{eq:Combined_Loss}). Figure \ref{fig:Re100_results} shows that the $\iLED$ model is able to accurately reconstruct the system state after multiple shedding cycles. Similarly to the Fitz-Hugh Nagomo case (sec \ref{sec:FHN}) the results underline the effectiveness of the $\iLED$ architecture, as the figure shows that most of the dynamics are supported by the linear part of the model. 

Finally, the natural frequency of $1.466 Hz$ learned by the \iLED linear operator is in accordance with the system data which presents a dominant frequency of $1.562 Hz$: This further confirms the validity of the model.

\begin{figure}[h]
    \centering
    \includegraphics[width=\textwidth]{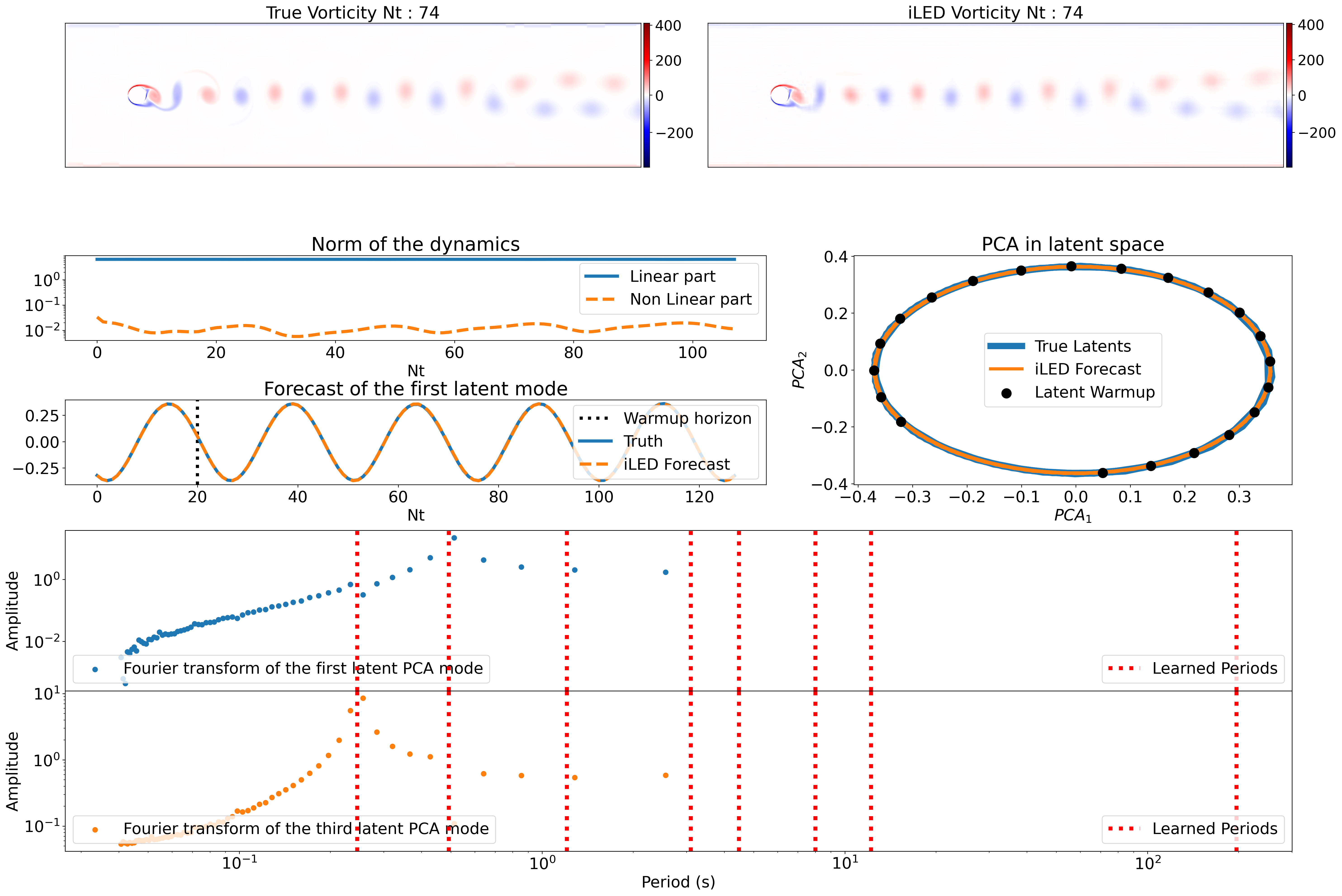}
    \caption{Results obtained with the iLED method on the case of the cylinder flow at a
Reynolds number of 750.}
    \label{fig:Re750_results}
\end{figure}

The results obtained on the case of the cylinder flow under a Reynolds number of 750 are presented in figure \ref{fig:Re750_results}. This case presents more complex dynamics than the simple 2D periodic limit cycle encountered for $\textit{Re}=100$. Following \cite{AdaLED}

We  used a latent dimension of $d_z=16$. Similar to the $\textit{Re}=100$ case, the \iLED model is able to accurately forecast and reconstruct the system state and once again, despite the higher complexity of the case, most of the dynamics are supported by the linear operator and the neural network closure ($\mathbf{\Psi}_1$ in equation \eqref{eq:iLED_ODE}) is only used to correct the numerical imperfections in the curvature of the learned latent attractor. Figure \ref{fig:Re750_results} also shows that the learned frequencies are coherent with the system data, as the two first natural frequencies of the linear operator are perfectly coherent with the dominant frequencies of the Fourier transform of the true latent trajectories. 

These results  demonstrate the ability of the \iLED model to capture the dynamics of two-dimensional bluff body flows up to Re=700. The model yields satisfying performance combined with a high degree of interpretability and stability.

\subsection{Remarks on the linearity of the dynamics}

In two of the three numerical experiments presented above, we have shown that the \iLED model was able to transform high dimensional, nonlinear PDEs into \textit{quasi}-linear Ordinary Differential Equations. This is in fact coherent with the oscillatory nature of the dynamics in both the FHN and Cylinder flow cases. We underline this result as a major strength of the \iLED framework as it is able to identify simple models from data without any \textit{a-priori} assumptions on the system under study. 

Moreover, it is important to note that although the identified models rely almost entirely on the linear part of the dynamics, they are still completed by a nonlinear term of lesser magnitude. Indeed, the complexity involved in learning a purely linear model reaching the same degree of accuracy might be higher. This is due to the fact that the neural autoencoders used for dimensionality reduction struggle to learn perfectly \textit{organized} latent attractors, which is critical to ensure the accuracy of purely linear dynamics. Of course, this aspect is only magnified with the increasing complexity of the application case. Thus, the nonlinear term in the framework can be looked at as a relaxation of the constraints on the shape of the latent attractor while still allowing for the extraction of a simple interpretable model, as the observed low magnitude of the nonlinear dynamics allows for accurate analysis of the model from the learned linear term.

Finally, we note that these \textit{quasi-}linear dynamics were not observed in the case of the Kuramoto-Shivasinsky equations. This is due to the chaotic nature of the case. Because the system does not rely on a set of clearly identified driving frequencies, the linear part of the model is not sufficient for the accurate representation of the case, and the non-linear part then automatically learns to complete the dynamics. This once again underlines the \textit{adaptability} of the model as no \textit{a-priori} knowledge of the nature of the dynamics is required to model the system.

\section{Conclusion} \label{Sec:conlcusions}

We present the \iLED method to learn interpretable reduced order dynamics for high-dimensional, multiscale systems. This method is closely related to by  Koopman operator theory and enahanced the Mori-Zwanzig formalism.  In addition to offering a high degree of interpretability, the latent dynamics of \iLED are  expressive enough such that the method can be applied to various problems.\\

We show that the approach performs well on a range of dynamics, from chaotic problems to high-dimensional 2D flow cases. For each test case, the method is able to learn a linear model for the latent dynamics as well as a non-Markovian, non-linear closure term. The high-dimensional systems are mapped with a non-linear encoder to a latent space, in which the complex non-linear PDEs can be reduced to very simple \textit{quasi}-linear ODEs, thus yielding fast and stable simulations. The high-dimensional space can be reconstructed from the latent space using a decoder that is trained simultaneously with the aforementioned encoder, using an autoencoder architecture.\\

Currently, the latent dynamics are computed deterministially. Future work, will address a  probabilistic version based on either the Bayesian approach or Conformal Inference, in order to quantify the uncertainty caused by dimensionality and model reduction. Another unsolved challenge pertains to the optimal choice of the latent dimension. As we discussed in section \ref{sec:training} various approaches can be used to estimate this value but the optimization of the latent dimension is not always feasible because of computational costs.\\
The \iLED method could be used to model more complex problems such as partially observed systems, or applied to real-world problems with unknown dynamics such as epidemic dynamics or brain activity to help deriving interpretable dynamical laws from available data.

\section*{Acknowledgement}
We would like to thank Sergey Litvinov for his help with generating the data for the flow around a cylinder example. Moreover, we thank Hunter Heidenreich, Michele Alessandro Bucci, and Lionel Mathelin for helpful discussions and feedback on the manuscript. E.M., M.Y. and M.S. acknowledge the support of the French government under the France 2030 program as part of the SystemX Technological Research Institute. S.K. and P.K. acknowledge support by The European High Performance Computing Joint Undertaking (EuroHPC) Grant
DCoMEX (956201-H2020-JTI-EuroHPC-2019-1).

\appendix
\section{Runge-Kutta}
\label{sec:AppB}
This appendix contains the siRK3 scheme that advances the state $[\boldsymbol{z}_0,h_0]$ for a time step $\Delta t$ as follows~:

\begin{align}
\begin{split}
    \left( I - \frac{k}{6}\Delta t \mbf{A}_\theta \right) \boldsymbol{z}_{k\Delta t/3} &= \boldsymbol{z}_0 + \frac{k}{6}\Delta t \mathbf{A}_\theta \boldsymbol{z}_0  + \frac{k}{3}\Delta t \mbf{\Psi}_{\theta,1}(\boldsymbol{z}_{(k-1)\Delta t/3},\boldsymbol{h}_{(k-1)\Delta t/3}),\\
    \left( I - \frac{k}{6}\Delta t \mbf{\Lambda}_\theta \right) \boldsymbol{h}_{k\Delta t/3} &= \boldsymbol{h}_0 + \frac{k}{6}\Delta t \mathbf{\Lambda}_\theta \boldsymbol{h}_0  + \frac{k}{3}\Delta t \mbf{\Psi}_{\theta,2}(\boldsymbol{z}_{(k-1)\Delta t/3}),\\
    k=1,2,3.
\end{split}
\end{align}

Here $\mathbf{\Lambda}_\theta,\mbf{\Psi}_{\theta,1},\mbf{\Psi}_{\theta,2}$ and $\mbf{A}_\theta$ are based upon the \iLED architecture introduced in \ref{eq:iLED}.

\section{Details on the loss}
\label{sec:Loss}
As mentioned in sec \ref{sec:training}, two loss terms are sufficient to train the \iLED architecture~:

\begin{equation}
    \mathcal{L} = \frac{1}{N_t} \sum_{i=1}^{N_t} \Bigg[\underbrace{\Vert \Phi^\star_{t_i} - \mathcal{D}(\mathcal{E}(\Phi^\star_{t_i})) \Vert_2^2}_{\mathcal{L}_\text{rec}} \,
    + \, \alpha_1 \underbrace{\Vert \hat{\boldsymbol{z}}_{t_i} - \mathcal{E}(\Phi^\star_{t_i}) \Vert_2^2}_{\mathcal{L}_\text{forecast}} \Bigg].
\end{equation}

Where $\mathbf{\Phi}^\star$ denotes the true system's states extracted from the training trajectories, $N_t$ the length of the trajectories, and $\hat{\boldsymbol{z}}$ the reduced states predicted by integrating the $\iLED$ model in time.

We also found that adding terms of lesser importance was beneficial and helped stabilize training. A reconstructed forecast loss~:

\begin{equation}
    \mathcal{L}_\textit{rec forecast} =  \frac{1}{N_t} \sum_{i=1}^{N_t} \Vert \mathbf{\Phi}^\star_{t_i} - \mathcal{D}(\hat{\boldsymbol{z}}_{t_i}) \Vert_2^2.
\end{equation}

And a regularization loss on the nonlinear part of the \iLED dynamics~:

\begin{equation}
    \mathcal{L}_\textit{non-linearity} =  \frac{1}{N_t} \sum_{i=1}^{N_t} \Vert \mathbf{\Psi}_{1}(\hat{\boldsymbol{z}}_{t_i},\hat{\boldsymbol{h}}_{t_i}) \Vert_2^2,
\end{equation}

where $\mathbf{\Psi}_{1}$ is the non linear part of the \iLED dynamics in equation \eqref{eq:iLED}. Finally, the full loss is written as follows~:

\begin{equation}
    \mathcal{L} = \mathcal{L}_\textit{rec} + \alpha_1 \mathcal{L}_\textit{forecast} + \alpha_2 \mathcal{L}_\textit{rec forecast} + \alpha_3 \mathcal{L}_\textit{non-linearity},
\end{equation}

with the coefficients $\alpha_i$ adjusted to control the importance of each term.
\section{Engineering Details}
\label{sec:eng_details}
This section discusses the choices we made during the creation of the method that should be considered to reproduce the results.

\subsection{Memory Initialization}

Thanks to the memory architecture of the \iLED model, the memory term $h$ can be initialized to an arbitrary degree of accuracy from the history of the solution. Indeed, the value of the memory at $t=0$ is computed as follows~:

\begin{equation}
\label{eq:infinite_memory}
    h_0 = \int_{-\infty}^{0} \mathbf{\Psi}_1 (\boldsymbol{z}(s)) e^{-\mathbf{\Lambda} s} ds.
\end{equation}

The infinite boundary of the above integral can be relaxed by computing the longest time horizon $\tau_{max}$ of the memory from the largest entry $\lambda_\textit{max}$ of the negative diagonal matrix $\mathbf{\Lambda}$~:

\begin{equation}
    \tau_\textit{max} = \frac{\epsilon}{\lambda_\textit{max}}
\end{equation}

Where $\epsilon \in \mathbb{R}^+$ is a small parameter, generally chosen to be equal to $10^{-2}$, that controls the relative error on the computation of $h_0$. After relaxing the infinite boundary in equation \eqref{eq:infinite_memory}, the memory can be initialised as follows~:

\begin{equation}
    h_0 = \int_{\tau_\textit{max}}^0  \mathbf{\Psi}_1 (\mathcal{E}(\mathbf{\Phi}^\star (s))) e^{-\mathbf{\Lambda} s} ds.
\end{equation}

Note that the above integral can be computed from the training data as a simple trapezoidal integration, which can be directly backpropagated through during training. 

\subsection{Linear Parameterization}

To ensure a higher degree of stability in the model. The linear operator $\mathbf{A}_\theta$ in the \iLED architecture is parameterized to be stable as follows~:

\begin{equation}
    \mathbf{A}_\theta = \mathbf{W}_\theta - \mathbf{W}_\theta^T - \textrm{diag}(\textrm{abs}(\Vec{\boldsymbol{w}}_\theta)),
\end{equation}

with $\mathbf{W} \in \mathbb{R}^{d_z \times d_z}$ a trainable weight matrix and $\Vec{\boldsymbol{w}}_\theta \in \mathbb{R}^{d_z}$ a trainable vector. With this formulation, the operator $\mathbf{A}_\theta$ is guaranteed to be stable \textit{i.e.} its eigenvalues have negative or zero real parts. This not only stabilizes the model but also avoids divergence of the model in the early stages of training.

\subsection{Latent space centering}

To allow for the interpretability of the linear term in the \iLED dynamics, it is important to ensure that the latent codes computed by the encoder $\mathcal{E}$ are centered. Indeed, a limit cycle arising from an unforced linear system will necessarily be centered around the origin. To do so, we define a \textit{LatentSpaceCentering} operation $\mathbf{LC}(z)$ as follows~:

\begin{equation}
    \mathbf{LC}(\boldsymbol{z}) = \boldsymbol{z} - \Vec{\mu}.
\end{equation}

Where $\Vec{\mu}$ is a running mean of the latent code's averages that is computed during training and frozen at inference time. This approach is very similar to classical batch normalization, except the data is only centered, as unitary scaling of the latent space is not required for the model to learn efficiently.

\subsection{Form of the memory kernel network}

The neural network $\mathbf{\Psi}_2$ in equation \ref{eq:iLED} is used to \textit{lift} the latent codes $\boldsymbol{z}$ to a space of arbitrary dimension. During training, it can learn to compute useful features of $\boldsymbol{z}$ that can then be integrated in time. To simplify the training, we propose a modification of the classical multi layer perceptron which we denote as \textit{AugmentedIdentityEncoder} in the manuscript.

With this modification, the latent code $\boldsymbol{z}$ is added to the prediction of the neural network since the latent code already holds useful information in itself~:

\begin{equation}\label{eq:AugmentedIdentity}
    \mathbf{\Psi}_2(\boldsymbol{z}) = [\boldsymbol{z},\mathcal{MLP}(\boldsymbol{z})].
\end{equation}

Where $\mathcal{MLP} : \mathbb{R}^{d_z}\mapsto \mathbb{R}^{d_h - d_z}$ denotes a standard multi layer perceptron.

\section{Network parameters}
This section lists the various hyperparameters and network architectures used to obtain the results presented in section \ref{sec:examples}.

\subsection{FHN}
\label{sec:params_fhn}
The tables below present the architecture of both the autoencoder and \iLED dynamical models used to obtain the results on the FHN case.

\begin{table}[H]
    \centering
    \begin{tabular}{|>{\centering}m{.09\textwidth}|m{.9\textwidth}<{\centering}|}
    \hline
       Layer  & Encoder \\
    \hline
    (1) & ConstantPad1d(padding=(13, 14), value=0.0)\\
    (2) & Conv1d(2, 8, kernel\_size=(5,), stride=(1,), padding=same)\\
    (3) & AvgPool1d(kernel\_size=(2,), stride=(2,), padding=(0,))\\
    (4) & SiLU()\\
    (5) & Conv1d(8, 16, kernel\_size=(5,), stride=(1,), padding=same)\\
    (6) & AvgPool1d(kernel\_size=(2,), stride=(2,), padding=(0,))\\
    (7) & SiLU()\\
    (8) & Conv1d(16, 32, kernel\_size=(5,), stride=(1,), padding=same)\\
    (9) & AvgPool1d(kernel\_size=(2,), stride=(2,), padding=(0,))\\
    (10) & SiLU()\\
    (11) & Conv1d(32, 4, kernel\_size=(5,), stride=(1,), padding=same)\\
    (12) & AvgPool1d(kernel\_size=(2,), stride=(2,), padding=(0,))\\
    (13) & SiLU()\\
    (14) & Flatten(start\_dim=-2, end\_dim=-1)\\
    (15) & Linear(in\_features=32, out\_features=2, bias=True)\\
    (16) & LatentSpaceCenteringLayer()\\
    \hline
    \end{tabular}
    \begin{tabular}{|>{\centering}m{.09\textwidth}|m{.9\textwidth}<{\centering}|}
    \hline
    Layer  & Decoder \\
    \hline
    (1) & Linear(in\_features=2, out\_features=32, bias=True)\\
    (2) & SiLU()\\
    (3) & Unflatten(dim=-1, unflattened\_size=(4, 8))\\
    (4) & Upsample(scale\_factor=2.0, mode=linear)\\
    (5) & ConvTranspose1d(4, 32, kernel\_size=(5,), stride=(1,), padding=(2,))\\
    (6) & SiLU()\\
    (7) & Upsample(scale\_factor=2.0, mode=linear)\\
    (8) & ConvTranspose1d(32, 16, kernel\_size=(5,), stride=(1,), padding=(2,))\\
    (9) & SiLU()\\
    (10) & Upsample(scale\_factor=2.0, mode=linear)\\
    (11) & ConvTranspose1d(16, 8, kernel\_size=(5,), stride=(1,), padding=(2,))\\
    (12) & SiLU()\\
    (13) & Upsample(scale\_factor=2.0, mode=linear)\\
    (14) & ConvTranspose1d(8, 2, kernel\_size=(5,), stride=(1,), padding=(2,))\\
    (15) & 1 + 0.5 Tanh()\\
    (16) & Unpad()\\
    \hline
    \end{tabular}
    
    \caption{One-dimensional convolutional autoencoder used to obtain the results on the case of the FHN model presented in section \ref{sec:FHN}}
    \label{tab:FHN_CNN}
\end{table}

\begin{table}[H]
    \centering
    \begin{tabular}{|>{\centering}m{.3\textwidth}|m{.69\textwidth}<{\centering}|}
        \hline
         & \iLED Parameters\\
         \hline
         $\mathbf{A}_\theta$ & Linear(2,2,bias=False)  \\
         $\mathbf{\Psi}_1$ neurons & 18 - 32 - 32 - 32 - 2\\
         $\mathbf{\Psi}_1$ activation & SiLU()\\
         $d_h$ & 16\\
         $\mathbf{\Psi_2}$ & AugmentedIdentityEncoder (see Eq. \eqref{eq:AugmentedIdentity})\\
         $\mathbf{\Psi_2}$ neurons & 2 - 5 - 8 - 11 - 14\\
         $\mathbf{\Psi_2}$ activation & SiLU()\\
         $\mathbf{\Lambda}_\theta$  & $\text{diag}(\boldsymbol{w}), \, \boldsymbol{w} \in \mathbb{R}_-^{d_h}$\\
         \hline
    \end{tabular}
    \caption{Hyperparameters of the \iLED dynamics used to obtain the results on the FHN case presented in section \ref{sec:FHN}.}
    \label{tab:FHN_Dyn}
\end{table}

\subsection{KS}
\label{sec:params_KS}
Similar to the previous paragraph, the architecture of the networks used for the KS case are presented below.

\begin{table}[H]
    \centering
    \begin{tabular}{|>{\centering}m{.09\textwidth}|m{.9\textwidth}<{\centering}|}
    \hline
       Layer  & Encoder \\
       \hline
       (1) & Conv1d(1, 16, kernel\_size=(5,), stride=(1,), padding=same)\\
       (2) & AvgPool1d(kernel\_size=(2,), stride=(2,), padding=(2), (0,))\\
       (3) & SiLU()\\
       (4) & Conv1d(16, 32, kernel\_size=(5,), stride=(1,), padding=same)\\
       (5) & AvgPool1d(kernel\_size=(2,), stride=(2,), padding=(0,))\\
       (6) & SiLU()\\
       (7) & Conv1d(32, 64, kernel\_size=(5,), stride=(1,), padding=same)\\
       (8) & AvgPool1d(kernel\_size=(2,), stride=(2,), padding=(0,))\\
       (9) & SiLU()\\
       (10) & Conv1d(64, 8, kernel\_size=(5,), stride=(1,), padding=same)\\
       (11) & AvgPool1d(kernel\_size=(2,), stride=(2,), padding=(0,))\\
       (12) & SiLU()\\
       (13) & Flatten(start\_dim=-2, end\_dim=-1)\\
       (14) & Linear(in\_features=64, out\_features=8, bias=True)\\
       (15) & LatentSpaceCentering()\\
    \hline
    \end{tabular}
    \begin{tabular}{|>{\centering}m{.09\textwidth}|m{.9\textwidth}<{\centering}|}
    \hline
       Layer  & Decoder \\
       \hline
       (1) & Linear(in\_features=8, out\_features=64, bias=True)\\
       (2) & Unflatten(dim=-1, unflattened\_size=(8, 8))\\
       (3) & Upsample(scale\_factor=2.0, mode=linear)\\
       (4) & ConvTranspose1d(8, 64, kernel\_size=(5,), stride=(1,), padding=(2,))\\
       (5) & SiLU()\\
       (6) & Upsample(scale\_factor=2.0, mode=linear)\\
       (7) & ConvTranspose1d(64, 32, kernel\_size=(5,), stride=(1,), padding=(2,))\\
       (8) & SiLU()\\
       (9) & Upsample(scale\_factor=2.0, mode=linear)\\
       (10) & ConvTranspose1d(32, 16, kernel\_size=(5,), stride=(1,), padding=(2,))\\
       (11) & SiLU()\\
       (12) & Upsample(scale\_factor=2.0, mode=linear)\\
       (13) & ConvTranspose1d(16, 1, kernel\_size=(5,), stride=(1,), padding=(2,))\\
    \hline
    \end{tabular}
    \caption{One-dimensional convolutional autoencoder used to obtain the results on the case of the KS equation (sec \ref{sec:KS})}
    \label{tab:KS_CNN}
\end{table}

\begin{table}[H]
    \centering
    \begin{tabular}{|>{\centering}m{.3\textwidth}|m{.69\textwidth}<{\centering}|}
        \hline
         & \iLED Parameters\\
         \hline
         $\mathbf{A}_\theta$ & $\mathbf{W} - \mathbf{W}^T - \text{diag}(\boldsymbol{w}), \, \mathbf{W} \in \mathbb{R}^{d_z \times d_z}, \boldsymbol{w} \in \mathbb{R}_+^{d_z}$  \\
         $\mathbf{\Psi}_1$ neurons & 40 - 64 - 64 - 64 - 8\\
         $\mathbf{\Psi}_1$ activation & SiLU()\\
         $d_h$ & 32\\
         $\mathbf{\Psi_2}$ & AugmentedIdentityEncoder (see Eq. \eqref{eq:AugmentedIdentity})\\
         $\mathbf{\Psi_2}$ neurons & 8 - 12 - 16 - 20 - 24\\
         $\mathbf{\Psi_2}$ activation & SiLU()\\
         $\mathbf{\Lambda}_\theta$  & $\text{diag}(\boldsymbol{w}), \, \boldsymbol{w} \in \mathbb{R}_-^{d_h}$\\
         \hline
    \end{tabular}
    \caption{Hyperparameters of the \iLED dynamics used to obtain the results on the KS case presented in section \ref{sec:KS}.}
    \label{tab:KS_Dyn}
\end{table}

\subsection{Flow around a cylinder}
\label{sec:params_flow}

 The autoencoders used in the Cylinder flow case have a complex architecture, to simplify the notation, we define two blocks that combine similar operations~:

\begin{table}[H]
    \caption{Sub blocks defined to help describe the CNN autoencoders}
    \centering
        \begin{tabular}{|>{\centering}m{.09\textwidth}|m{.9\textwidth}<{\centering}|}
            \hline
            Layer & DownBlock(in\_size,out\_size)\\
            \hline
            (1) & Conv2d(in\_size, out\_size, kernel\_size=(5, 5), stride=(2, 2), padding=(2, 2), padding\_mode=replicate)\\
            (2) & SiLU()\\
            \hline
        \end{tabular}
        
        \begin{tabular}{|>{\centering}m{.09\textwidth}|m{.9\textwidth}<{\centering}|}
            \hline
            Layer & UpBlock(in\_size,out\_size)\\
            \hline
            (1) & Upsample(scale\_factor=2.0, mode=bilinear)\\
            (2) & Conv2d(in\_size, out\_size, kernel\_size=(5, 5), stride=(1, 1), padding=(2, 2), padding\_mode=replicate)\\
            (3) & SiLU()\\
            (4) & BatchNorm2d()\\
            \hline
        \end{tabular}
\end{table}

\begin{table}[H]
    \centering
    \begin{tabular}{|>{\centering}m{.09\textwidth}|m{.449\textwidth}<{\centering}|m{.449\textwidth}<{\centering}|}
    \hline
       Layer  & Encoder \#1 & Encoder \#2 \\
    \hline
       (1) & DownBlock(2,4) & DownBlock(2,4)\\
       (2) & DownBlock(4,16) & DownBlock(4,8)\\
       (3) & DownBlock(16,4) & DownBlock(8,16)\\
       (4) & DownBlock(4,2) & DownBlock(16,2)\\
       (5) & Flatten(start=-3,end=-1) & Flatten(start=-3,end=-1)\\
       (6) & Linear(512,20) & Linear(200,20)\\
       (7) & $d_{z_1} = 20$ & $d_{z_2} = 20$\\
       \hline
        & \multicolumn{2}{c|}{Mixer Encoder}\\
       \hline
       (8) & \multicolumn{2}{c|}{Concatenate($z_1,z_2$)}\\
       (9) & \multicolumn{2}{c|}{Linear(40,30)}\\
       (10) & \multicolumn{2}{c|}{SiLU()}\\
       (11) & \multicolumn{2}{c|}{Linear(30,$d_z$)}\\
    \hline
    \end{tabular}
    \begin{tabular}{|>{\centering}m{.09\textwidth}|m{.449\textwidth}<{\centering}|m{.449\textwidth}<{\centering}|}
    
    \hline
    Layer & \multicolumn{2}{c|}{Mixer Decoder}\\
       \hline
       (1) & \multicolumn{2}{c|}{Linear($d_z$,30)}\\
       (2) & \multicolumn{2}{c|}{SiLU()}\\
       (3) & \multicolumn{2}{c|}{Linear(30,40)}\\
       (4) & \multicolumn{2}{c|}{$z_1, z_2=z$}\\
    \hline
         & Decoder \#1 & Decoder \#2 \\
    \hline
       (5) & Linear(20,512) & Linear(20,200)\\
       (6) & Unflatten(-1,(2,32,8)) & Unflatten(-1,(2,10,10)\\
       (7) & UpBlock(2,4) & UpBlock(2,16)\\
       (8) & UpBlock(4,16) & UpBlock(16,8)\\
       (9) & UpBlock(16,4) & UpBlock(8,4)\\
       (10) & Upsample(2.0,bilinear) & Upsample(2.0,bilinear) \\
       (11) & Conv2d(4, 1, kernel\_size=(5, 5), stride=(1, 1), padding=(2, 2), padding\_mode=replicate) & Conv2d(4, 1, kernel\_size=(5, 5), stride=(1, 1), padding=(2, 2), padding\_mode=replicate)\\
       (12) & Flatten(start=-3,end=-1) & Flatten(start=-3,end=-1)\\
       (13) & StreamFnToVelocity() & StreamFnToVelocity()\\
       \hline
       
    \end{tabular}
    \caption{Hyperparameters of the 2-dimensional convolutional autoencoder used to obtain the results on the Cylinder case presented in section \ref{sec:Cylinder}.}
    \label{tab:Cyl_CNN}
\end{table}

The value of $d_z$ changes depending on the Reynolds number considered. It is equal to $d_z=3$ in the $Re=100$ case, and $d_z = 16$ in the $Re=750$ case. Note that we don't use a \textit{LatentSpaceCentering} layer contrary to the other cases. This is due to the fact that, because of the memory costs of the models, the batch size has to be relatively low, which has a negative impact on batch normalization approaches. To ensure that the latent space remained centered, which is critical to the accuracy and interpretability of the linear term in the dynamics, we added a term to the loss~:

\begin{equation}\label{eq:centering_loss}
    \mathcal{L}_{centering} = \Vert \frac{1}{N_T} \sum_{i=1}^{N_T} \mathcal{E}(\mathbf{\Phi}_{t_i})) \Vert_2^2.
\end{equation}

This loss effectively penalizes the average of the latent codes, ensuring that they are centered around the origin.

Also note that the decoder doesn't directly predict the velocity field, but the stream function $\psi$ which is a scalar field, that is used to compute the velocity components as follows~:

\begin{equation}
    u_x = \pdv{\psi}{y}, \: u_y = -\pdv{\psi}{x}.
\end{equation}

This specific choice is inspired from previous works (\cite{mohan2020embedding,AdaLED}). It allows for the guaranteed prediction of a divergence free field, which is a constraint of the incompressible Navier-Stokes equations. 

Finally, the parameters of the \iLED dynamical model are provided below~:

\begin{table}[H]
    \centering
    \begin{tabular}{|>{\centering}m{.3\textwidth}|m{.69\textwidth}<{\centering}|}
        \hline
         & \iLED Parameters\\
         \hline
         $\mathbf{A}_\theta$ & $\mathbf{W} - \mathbf{W}^T - \text{diag}(\boldsymbol{w}), \, \mathbf{W} \in \mathbb{R}^{d_z \times d_z}, \boldsymbol{w} \in \mathbb{R}_+^{d_z}$  \\
         $\mathbf{\Psi}_1$ neurons & $d_z+d_h$ - 128 - 128 - 128 - $d_z$\\
         $\mathbf{\Psi}_1$ activation & SiLU()\\
         $d_h$ & 32\\
         $\mathbf{\Psi_2}$ & AugmentedIdentityEncoder (see Eq. \eqref{eq:AugmentedIdentity})\\
         $\mathbf{\Psi_2}$ neurons ($Re=750$)& 16 - 17 - 17 - 17 - 15\\
         $\mathbf{\Psi_2}$ neurons ($Re=100$)& 3 - 10 - 16 - 22 - 28\\
         $\mathbf{\Psi_2}$ activation & SiLU()\\
         $\mathbf{\Lambda}_\theta$  & $\text{diag}(\boldsymbol{w}), \, \boldsymbol{w} \in \mathbb{R}_-^{d_h}$\\
         \hline
    \end{tabular}
    \caption{Hyperparameters of the \iLED dynamics used to obtain the results on the Cylinder cases presented in section \ref{sec:Cylinder}.}
    \label{tab:Cyl_Dyn}
\end{table}

 \bibliographystyle{elsarticle-num} 
 \bibliography{bib}

\begin{thebibliography}{10}
\expandafter\ifx\csname url\endcsname\relax
  \def\url#1{\texttt{#1}}\fi
\expandafter\ifx\csname urlprefix\endcsname\relax\def\urlprefix{URL }\fi
\expandafter\ifx\csname href\endcsname\relax
  \def\href#1#2{#2} \def\path#1{#1}\fi

\bibitem{wilcox1988multiscale}
D.~C. Wilcox, Multiscale model for turbulent flows, AIAA journal 26~(11) (1988) 1311--1320.

\bibitem{dura2019netpyne}
S.~Dura-Bernal, B.~A. Suter, P.~Gleeson, M.~Cantarelli, A.~Quintana, F.~Rodriguez, D.~J. Kedziora, G.~L. Chadderdon, C.~C. Kerr, S.~A. Neymotin, et~al., Netpyne, a tool for data-driven multiscale modeling of brain circuits, Elife 8 (2019) e44494.

\bibitem{climatenas}
N.~R. Council, A National Strategy for Advancing Climate Modeling, The National Academies Press, 2012.

\bibitem{mahadevan2016impact}
A.~Mahadevan, The impact of submesoscale physics on primary productivity of plankton, Annual review of marine science 8 (2016) 161--184.

\bibitem{Palmer2015}
T.~Palmer, \href{https://doi.org/10.1038/526032a}{Modelling: Build imprecise supercomputers}, Nature 526~(7571) (2015) 32--33.
\newblock \href {https://doi.org/10.1038/526032a} {\path{doi:10.1038/526032a}}.
\newline\urlprefix\url{https://doi.org/10.1038/526032a}

\bibitem{kevrekidis2004equation}
I.~G. Kevrekidis, C.~W. Gear, G.~Hummer, Equation-free: The computer-aided analysis of complex multiscale systems, AIChE Journal 50~(7) (2004) 1346--1355.

\bibitem{kevrekidis2003equation}
I.~G. Kevrekidis, C.~W. Gear, J.~M. Hyman, P.~G. Kevrekidis, O.~Runborg, C.~Theodoropoulos, et~al., Equation-free, coarse-grained multiscale computation: enabling microscopic simulators to perform system-level analysis, Commun. Math. Sci 1~(4) (2003) 715--762.

\bibitem{bar2019learning}
Y.~Bar-Sinai, S.~Hoyer, J.~Hickey, M.~P. Brenner, Learning data-driven discretizations for partial differential equations, Proceedings of the National Academy of Sciences 116~(31) (2019) 15344--15349.

\bibitem{weinan2003heterognous}
E.~Weinan, B.~Engquist, Z.~Huang, Heterogeneous multiscale method: a general methodology for multiscale modeling, Physical Review B 67~(9) (2003) 092101.

\bibitem{weinan2007heterogeneous}
E.~Weinan, B.~Engquist, X.~Li, W.~Ren, E.~Vanden-Eijnden, Heterogeneous multiscale methods: a review, Communications in computational physics 2~(3) (2007) 367--450.

\bibitem{tao2010nonintrusive}
M.~Tao, H.~Owhadi, J.~E. Marsden, Nonintrusive and structure preserving multiscale integration of stiff odes, sdes, and hamiltonian systems with hidden slow dynamics via flow averaging, Multiscale Modeling \& Simulation 8~(4) (2010) 1269--1324.

\bibitem{givon_extracting_2004}
D.~Givon, R.~Kupferman, A.~Stuart, Extracting {Macroscopic} {Dynamics}: {Model} {Problems} and {Algorithms}, Nonlinearity (2004).

\bibitem{mezic2005spectral}
I.~Mezi{\'c}, Spectral properties of dynamical systems, model reduction and decompositions, Nonlinear Dynamics 41 (2005) 309--325.

\bibitem{grigo2019physics}
C.~Grigo, P.-S. Koutsourelakis, A physics-aware, probabilistic machine learning framework for coarse-graining high-dimensional systems in the small data regime, Journal of Computational Physics 397 (2019) 108842.

\bibitem{kaltenbach2023interpretable}
S.~Kaltenbach, P.-S. Koutsourelakis, P.~Koumoutsakos, Interpretable reduced-order modeling with time-scale separation, arXiv preprint arXiv:2303.02189 (2023).

\bibitem{robinson1994inertial}
J.~C. Robinson, Inertial manifolds for the kuramoto-sivashinsky equation, Physics Letters A 184~(2) (1994) 190--193.

\bibitem{RowleyReview}
C.~W. Rowley, S.~T. Dawson, \href{https://doi.org/10.1146/annurev-fluid-010816-060042}{Model reduction for flow analysis and control}, Annual Review of Fluid Mechanics 49~(1) (2017) 387--417.
\newblock \href {http://arxiv.org/abs/https://doi.org/10.1146/annurev-fluid-010816-060042} {\path{arXiv:https://doi.org/10.1146/annurev-fluid-010816-060042}}, \href {https://doi.org/10.1146/annurev-fluid-010816-060042} {\path{doi:10.1146/annurev-fluid-010816-060042}}.
\newline\urlprefix\url{https://doi.org/10.1146/annurev-fluid-010816-060042}

\bibitem{PEHERSTORFER201521}
B.~Peherstorfer, K.~Willcox, \href{https://www.sciencedirect.com/science/article/pii/S0045782515001280}{Dynamic data-driven reduced-order models}, Computer Methods in Applied Mechanics and Engineering 291 (2015) 21--41.
\newblock \href {https://doi.org/https://doi.org/10.1016/j.cma.2015.03.018} {\path{doi:https://doi.org/10.1016/j.cma.2015.03.018}}.
\newline\urlprefix\url{https://www.sciencedirect.com/science/article/pii/S0045782515001280}

\bibitem{choi2019space}
Y.~Choi, K.~Carlberg, Space--time least-squares petrov--galerkin projection for nonlinear model reduction, SIAM Journal on Scientific Computing 41~(1) (2019) A26--A58.

\bibitem{CD-ROM}
E.~Menier, M.~A. Bucci, M.~Yagoubi, L.~Mathelin, M.~Schoenauer, \href{https://arxiv.org/abs/2202.10746}{Cd-rom: Complementary deep-reduced order model} (2022).
\newblock \href {https://doi.org/10.48550/ARXIV.2202.10746} {\path{doi:10.48550/ARXIV.2202.10746}}.
\newline\urlprefix\url{https://arxiv.org/abs/2202.10746}

\bibitem{lee2020model}
K.~Lee, K.~T. Carlberg, Model reduction of dynamical systems on nonlinear manifolds using deep convolutional autoencoders, Journal of Computational Physics 404 (2020) 108973.

\bibitem{vlachas2022multiscale}
P.~R. Vlachas, G.~Arampatzis, C.~Uhler, P.~Koumoutsakos, Multiscale simulations of complex systems by learning their effective dynamics, Nature Machine Intelligence 4~(4) (2022) 359--366.

\bibitem{hochreiter1997long}
S.~Hochreiter, J.~Schmidhuber, Long short-term memory, Neural computation 9~(8) (1997) 1735--1780.

\bibitem{mori1965transport}
H.~Mori, Transport, collective motion, and brownian motion, Progress of theoretical physics 33~(3) (1965) 423--455.

\bibitem{zwanzig1973nonlinear}
R.~Zwanzig, Nonlinear generalized langevin equations, Journal of Statistical Physics 9~(3) (1973) 215--220.

\bibitem{koopman_hamiltonian_1931}
B.~O. Koopman, \href{https://www.jstor.org/stable/86114}{Hamiltonian {Systems} and {Transformations} in {Hilbert} {Space}}, Proceedings of the National Academy of Sciences of the United States of America 17~(5) (1931) 315--318.
\newline\urlprefix\url{https://www.jstor.org/stable/86114}

\bibitem{mezic2013analysis}
I.~Mezi{\'c}, Analysis of fluid flows via spectral properties of the koopman operator, Annual review of fluid mechanics 45 (2013) 357--378.

\bibitem{chen2018neural}
T.~Q. Chen, Y.~Rubanova, J.~Bettencourt, D.~K. Duvenaud, Neural ordinary differential equations, in: Advances in neural information processing systems, 2018, pp. 6571--6583.

\bibitem{li2020scalable}
X.~Li, T.-K.~L. Wong, R.~T. Chen, D.~Duvenaud, Scalable gradients for stochastic differential equations, arXiv preprint arXiv:2001.01328 (2020).

\bibitem{champion2019discovery}
K.~P. Champion, S.~L. Brunton, J.~N. Kutz, Discovery of nonlinear multiscale systems: Sampling strategies and embeddings, SIAM Journal on Applied Dynamical Systems 18~(1) (2019) 312--333.

\bibitem{lusch2018deep}
B.~Lusch, J.~N. Kutz, S.~L. Brunton, Deep learning for universal linear embeddings of nonlinear dynamics, Nature communications 9~(1) (2018) 1--10.

\bibitem{schmid2010dynamic}
P.~J. Schmid, Dynamic mode decomposition of numerical and experimental data, Journal of fluid mechanics 656 (2010) 5--28.

\bibitem{williams2015data}
M.~O. Williams, I.~G. Kevrekidis, C.~W. Rowley, A data--driven approximation of the koopman operator: Extending dynamic mode decomposition, Journal of Nonlinear Science 25 (2015) 1307--1346.

\bibitem{doshi2017towards}
F.~Doshi-Velez, B.~Kim, Towards a rigorous science of interpretable machine learning, arXiv preprint arXiv:1702.08608 (2017).

\bibitem{koutnik2014clockwork}
J.~Koutnik, K.~Greff, F.~Gomez, J.~Schmidhuber, A clockwork rnn, in: International conference on machine learning, PMLR, 2014, pp. 1863--1871.

\bibitem{kaltenbach2020incorporating}
S.~Kaltenbach, P.-S. Koutsourelakis, Incorporating physical constraints in a deep probabilistic machine learning framework for coarse-graining dynamical systems, Journal of Computational Physics 419 (2020) 109673.

\bibitem{kaltenbach_physics-aware_2021-1}
S.~Kaltenbach, P.~S. Koutsourelakis, \href{https://openreview.net/forum?id=vyY0jnWG-tK}{Physics-aware, probabilistic model order reduction with guaranteed stability}, in: {International Conference on Learning Representations (ICLR)}, 2021.
\newline\urlprefix\url{https://openreview.net/forum?id=vyY0jnWG-tK}

\bibitem{Novati2021}
G.~Novati, H.~L. de~Laroussilhe, P.~Koumoutsakos, \href{https://doi.org/10.1038/s42256-020-00272-0}{Automating turbulence modelling by multi-agent reinforcement learning}, Nat. Mach. Intell. 3~(1) (2021) 87--96.
\newblock \href {https://doi.org/10.1038/s42256-020-00272-0} {\path{doi:10.1038/s42256-020-00272-0}}.
\newline\urlprefix\url{https://doi.org/10.1038/s42256-020-00272-0}

\bibitem{Amsallem2011}
D.~Amsallem, C.~Farhat, \href{https://doi.org/10.1137/100813051}{An online method for interpolating linear parametric reduced-order models}, SIAM Journal on Scientific Computing 33~(5) (2011) 2169--2198.
\newblock \href {http://arxiv.org/abs/https://doi.org/10.1137/100813051} {\path{arXiv:https://doi.org/10.1137/100813051}}, \href {https://doi.org/10.1137/100813051} {\path{doi:10.1137/100813051}}.
\newline\urlprefix\url{https://doi.org/10.1137/100813051}

\bibitem{GEELEN2023115717}
R.~Geelen, S.~Wright, K.~Willcox, \href{https://www.sciencedirect.com/science/article/pii/S0045782522006727}{Operator inference for non-intrusive model reduction with quadratic manifolds}, Computer Methods in Applied Mechanics and Engineering 403 (2023) 115717.
\newblock \href {https://doi.org/https://doi.org/10.1016/j.cma.2022.115717} {\path{doi:https://doi.org/10.1016/j.cma.2022.115717}}.
\newline\urlprefix\url{https://www.sciencedirect.com/science/article/pii/S0045782522006727}

\bibitem{girin2020dynamical}
L.~Girin, S.~Leglaive, X.~Bie, J.~Diard, T.~Hueber, X.~Alameda-Pineda, Dynamical variational autoencoders: A comprehensive review, arXiv preprint arXiv:2008.12595 (2020).

\bibitem{srivastava2015unsupervised}
N.~Srivastava, E.~Mansimov, R.~Salakhudinov, Unsupervised learning of video representations using lstms, in: International conference on machine learning, PMLR, 2015, pp. 843--852.

\bibitem{brunton2021modern}
S.~L. Brunton, M.~Budi{\v{s}}i{\'c}, E.~Kaiser, J.~N. Kutz, Modern koopman theory for dynamical systems, arXiv preprint arXiv:2102.12086 (2021).

\bibitem{budivsic2012applied}
M.~Budi{\v{s}}i{\'c}, R.~Mohr, I.~Mezi{\'c}, Applied koopmanism, Chaos: An Interdisciplinary Journal of Nonlinear Science 22~(4) (2012) 047510.

\bibitem{lin2021datadrivenmz}
Y.~T. Lin, Y.~Tian, M.~Anghel, D.~Livescu, Data-driven learning for the mori-zwanzig formalism: a generalization of the koopman learning framework (2021).

\bibitem{chorin2007problem}
A.~Chorin, P.~Stinis, Problem reduction, renormalization, and memory, Communications in Applied Mathematics and Computational Science 1~(1) (2007) 1--27.

\bibitem{darve2009computing}
E.~Darve, J.~Solomon, A.~Kia, Computing generalized langevin equations and generalized fokker--planck equations, Proceedings of the National Academy of Sciences 106~(27) (2009) 10884--10889.

\bibitem{kondrashov2015data}
D.~Kondrashov, M.~D. Chekroun, M.~Ghil, Data-driven non-markovian closure models, Physica D: Nonlinear Phenomena 297 (2015) 33--55.

\bibitem{chen1995universal}
T.~Chen, H.~Chen, Universal approximation to nonlinear operators by neural networks with arbitrary activation functions and its application to dynamical systems, IEEE Transactions on Neural Networks 6~(4) (1995) 911--917.

\bibitem{lu2021learning}
L.~Lu, P.~Jin, G.~Pang, Z.~Zhang, G.~E. Karniadakis, Learning nonlinear operators via deeponet based on the universal approximation theorem of operators, Nature machine intelligence 3~(3) (2021) 218--229.

\bibitem{kissas2022learning}
G.~Kissas, J.~H. Seidman, L.~F. Guilhoto, V.~M. Preciado, G.~J. Pappas, P.~Perdikaris, Learning operators with coupled attention, Journal of Machine Learning Research 23~(215) (2022) 1--63.

\bibitem{li2021fourier}
Z.~Li, N.~B. Kovachki, K.~Azizzadenesheli, B.~liu, K.~Bhattacharya, A.~Stuart, A.~Anandkumar, \href{https://openreview.net/forum?id=c8P9NQVtmnO}{Fourier neural operator for parametric partial differential equations}, in: International Conference on Learning Representations, 2021.
\newline\urlprefix\url{https://openreview.net/forum?id=c8P9NQVtmnO}

\bibitem{bollt2017edmd}
Q.~Li, F.~Dietrich, E.~M. Bollt, I.~G. Kevrekidis, \href{http://dx.doi.org/10.1063/1.4993854}{Extended dynamic mode decomposition with dictionary learning: A data-driven adaptive spectral decomposition of the koopman operator}, Chaos: An Interdisciplinary Journal of Nonlinear Science 27~(10) (2017) 103111.
\newblock \href {https://doi.org/10.1063/1.4993854} {\path{doi:10.1063/1.4993854}}.
\newline\urlprefix\url{http://dx.doi.org/10.1063/1.4993854}

\bibitem{otto2019lran}
S.~Otto, C.~Rowley, Linearly-recurrent autoencoder networks for learning dynamics, SIAM Journal on Applied Dynamical Systems 18 (12 2017).
\newblock \href {https://doi.org/10.1137/18M1177846} {\path{doi:10.1137/18M1177846}}.

\bibitem{nolitsa}
M.~Mannattil, Nolitsa (nonlinear time series analysis), \url{https://github.com/manu-mannattil/nolitsa} (2023).

\bibitem{Falconner2003fractal}
K.~Falconer, Fractal Geometry: Mathematical Foundations and Applications, 2003.
\newblock \href {https://doi.org/10.1002/0470013850} {\path{doi:10.1002/0470013850}}.

\bibitem{kar2006sirk3}
S.~K. Kar, A semi-implicit runge–kutta time-difference scheme for the two-dimensional shallow-water equations, Monthly Weather Review 134~(10) (2006) 2916--2926.
\newblock \href {https://doi.org/10.1175/MWR3214.1} {\path{doi:10.1175/MWR3214.1}}.

\bibitem{rocsoreanu2012fitzhugh}
C.~Rocsoreanu, A.~Georgescu, N.~Giurgiteanu, The FitzHugh-Nagumo model: bifurcation and dynamics, Vol.~10, Springer Science \& Business Media, 2012.

\bibitem{LB}
K.~I.~V., S.~Ansumali, F.~C.~E., C.~S.~S., \href{http://global-sci.org/intro/article_detail/cicp/7972.html}{Elements of the lattice boltzmann method i: Linear advection equation}, Communications in Computational Physics 1~(4) (2006) 616--655.
\newblock \href {https://doi.org/https://doi.org/} {\path{doi:https://doi.org/}}.
\newline\urlprefix\url{http://global-sci.org/intro/article_detail/cicp/7972.html}

\bibitem{hyman1986kuramoto}
J.~M. Hyman, B.~Nicolaenko, The kuramoto-sivashinsky equation: a bridge between pde's and dynamical systems, Physica D: Nonlinear Phenomena 18~(1-3) (1986) 113--126.

\bibitem{alessandro2022control}
M.~Bucci, O.~Semeraro, A.~Allauzen, L.~Cordier, L.~Mathelin, Nonlinear Optimal Control Using Deep Reinforcement Learning, 2022, pp. 279--290.
\newblock \href {https://doi.org/10.1007/978-3-030-67902-6_24} {\path{doi:10.1007/978-3-030-67902-6_24}}.

\bibitem{callaham2019robust}
J.~L. Callaham, K.~Maeda, S.~L. Brunton, Robust flow reconstruction from limited measurements via sparse representation, Physical Review Fluids 4~(10) (2019) 103907.

\bibitem{popinet2020basilisk}
S.~Popinet, Basilisk flow solver and pde library.

\bibitem{AdaLED}
I.~Kičić, P.~R. Vlachas, G.~Arampatzis, M.~Chatzimanolakis, L.~Guibas, P.~Koumoutsakos, Adaptive learning of effective dynamics: Adaptive real-time, online modeling for complex systems (2023).
\newblock \href {http://arxiv.org/abs/2304.01732} {\path{arXiv:2304.01732}}.

\bibitem{mohan2020embedding}
A.~T. Mohan, N.~Lubbers, D.~Livescu, M.~Chertkov, Embedding hard physical constraints in neural network coarse-graining of 3d turbulence (2020).
\newblock \href {http://arxiv.org/abs/2002.00021} {\path{arXiv:2002.00021}}.

\end{thebibliography}





\end{document}